\documentclass{article}
\usepackage{PRIMEarxiv}

\usepackage{graphicx}
\usepackage[english]{babel}
\usepackage{times} % Font
\usepackage{url}
\usepackage{verbatim}
\usepackage{mwe}
\usepackage{color}
\usepackage{hyperref}
\usepackage{amsmath}
\usepackage{bm}
\usepackage{multirow}
\usepackage[numbers]{natbib}
\usepackage{amsfonts} 

\usepackage{float}
\usepackage{mathtools}
\usepackage{enumitem}
\usepackage[ruled,vlined,linesnumbered,noend]{algorithm2e}
\usepackage[T1]{fontenc}

\DeclareMathOperator*{\argmax}{argmax} % no space, limits 

\usepackage{times}
\usepackage[utf8]{inputenc}

\usepackage{tabularx} 
\usepackage{ulem}
\usepackage[ruled,vlined,linesnumbered,noend]{algorithm2e}

\usepackage{caption}
\usepackage{subcaption}

\usepackage[misc]{ifsym}

\usepackage{xcolor}
\usepackage[colorinlistoftodos,prependcaption,textsize=tiny]{todonotes}

\emergencystretch=\maxdimen
\title{It's All in the Embedding! Fake News Detection Using Document Embeddings}

\author{
    Ciprian-Octavian~Truică$^{1}$,
    Elena-Simona~Apostol$^{1}$ \\
    $^1$ Computer Science and Engineering Department, Faculty of Automatic Control and Computers, \\ University Politehnica of Bucharest, Bucharest, Romania \\
  \texttt{ciprian.truica@upb.ro, elena.apostol@upb.ro}
}

\begin{document}

\maketitle              % typeset the header of the 

\begin{abstract}
With the current shift in the mass media landscape from journalistic rigor to social media,  personalized social media is becoming the new norm. Although the digitalization progress of the media brings many advantages, it also increases the risk of spreading disinformation, misinformation, and malformation through the use of fake news. The emergence of this harmful phenomenon has managed to polarize society and manipulate public opinion on particular topics, e.g., elections, vaccinations, etc. Such information propagated on social media can distort public perceptions and generate social unrest while lacking the rigor of traditional journalism. Natural Language Processing and Machine Learning techniques are essential for developing efficient tools that can detect fake news. Models that use the context of textual data are essential for resolving the fake news detection problem, as they manage to encode linguistic features within the vector representation of words. In this paper, we propose a new approach that uses document embeddings to build multiple models that accurately label news articles as reliable or fake. We also present a benchmark on different architectures that detect fake news using binary or multi-labeled classification. We evaluated the models on five large news corpora using accuracy, precision, and recall. We obtained better results than more complex state-of-the-art Deep Neural Network models. We observe that the most important factor for obtaining high accuracy is the document encoding, not the classification model's complexity.

% Keywords
\end{abstract}

\section{Introduction}
With the increase in the digitalization of mass media, new journalistic paradigms for information distribution have emerged.
These new paradigms have substantially changed the way society consumes information.
By trying to be ahead of the competition, sometimes people who report on world events leave behind the rigors of classical journalism and publish their content as soon as possible in order to "go viral" by obtaining as many views, likes, comments, and shares as possible in a short amount of time~\cite{Truica2021c}. 
This new paradigm centers on the users, catering to their needs, behavior, and interests.
Along with the advantages the digitalization of mass media brings, it also increases the risk of misinformation, with potentially detrimental consequences for society, by facilitating the spread of misinformation~\citep{Mustafaraj2017,Ruths2019} in the form of fake news (which influenced the Brexit referendum~\citep{Bastos2017}, the 2016 US presidential election~\citep{Bovet2019}, COVID-19 vaccinations~\citep{Rzymski2021}, etc.).

Fake news consists of news articles that are intentionally and verifiably false.
This type of information aims to mislead readers by presenting alleged, real-seeming facts about social, economic, and political subjects of interest~\citep{Truica2022b}.
However, the current technological trends make this type of content harmful, with potentially dire consequences to the community (e.g., public polarization regarding elections).
This has become a major challenge for democracy.
Information propagated online may lack the rigor of classic journalism, and can, therefore, distort public perceptions, cause false alarms, and generate social unrest.
Furthermore, the president of the EU, Ursula von der Leyen, has repeatedly condemned and asked for immediate action to be taken against the spread of fake news that undermines democracy and public health~\citep{EU2022}.
Thus, the ideological polarization of readers through the spread of fake news is an important issue and requires scholarly attention. 
We believe that designing and building tools and methods for accurately detecting fake news is of great relevance, and thus, our results will have an overall positive impact.

In this paper, we propose a new approach that uses document embeddings (i.e., \textsc{DocEmb}) for detecting fake news.
We also present our benchmark on different architectures that detect fake news using binary or multi-labeled classification.
The document embeddings are constructed using several (1) word embeddings trained on each dataset selected for experiments, and (2) pre-trained transformers. 
We employ TFIDF, word embeddings (i.e., \textsc{Word2Vec}, \textsc{FastText}, and \textsc{GloVe}), and transformers (i.e., BERT, \textsc{RoBERTa}, and BART) to create \textsc{DocEmb}, our new document embedding approach.
For classification, we train both classical Machine Learning models (i.e., Naïve Bayes, Gradient Boosted Trees) and Deep Learning models (i.e., Perceptron, Multi-Layer Perceptron, [Bi]LSTM, [Bi]GRU).

In our experiments, we analyze the performance of the \textsc{DocEmb} based detection solution on multiple datasets annotated with either binary or multi-class labels. 
We use 4 binary datasets, i.e., a sample of 20\,000 manually annotated news articles from the Fake News Corpus, Liar, Kaggle, and Buzz Feed News.
Finally, we use 2 multi-labeled datasets, i.e., Liar with 6 labels and TSHP-17 with 3 labels. 
As evaluation metrics, we use accuracy, precision, and recall.

We compare our results with state-of-the-art Deep Neural Networks models.
Our method outperforms these models on each dataset.
The most important takeaway from our experiments is that we empirically show that:
\begin{itemize}
    \item[\textit{(1)}] A simpler neural architecture offers better or at least similar results compared to complex architectures that employ multiple layers, and
    \item[\textit{(2)}] The difference in performance lies in the embeddings used to vectorize the textual data and how well these perform in encoding contextual and linguistic features. 
\end{itemize}

The main contributions of this article are as follows:
\begin{itemize}
    \item[$(C_1)$] We propose a new document embedding (\textsc{DocEmb}) constructed using word embeddings and transformers. We specifically trained the proposed \textsc{DocEmb} on the five datasets used in the experiments. 
    \item[$(C_2)$] We show empirically that simple Machine Leaning algorithms trained with our proposed \textsc{DocEmb} obtain similar results or better results than deep learning architectures specifically developed for the task of binary and multi-class fake news detection. This contribution is important in the machine learning literature because it changes the focus from the classification architecture to the document encoding architecture. 
    \item[$(C_3)$] We present a new manually filtered dataset. The original dataset is the widely used Fake News Corpus that was annotated with an automatic process.
\end{itemize}

This paper is structured as follows.
Section~\ref{sec:related_work} discusses current research on the topic of fake news detection. 
Section~\ref{sec:methodology} introduces our approach and presents the different modules and models employed.
Section~\ref{sec:results} presents the datasets and analyzes the results.
In Section~\ref{sec:discussion}, we summarize our key findings and discuss the major implications.
Section~\ref{sec:conclusions} presents the conclusions and outlines directions for future work.

\section{Related Work}\label{sec:related_work}

As views and clicks monetize online media, for some publishers, it is most important to provide news that might interest their audience, to the detriment of the quality of the facts reported~\citep{Chen2015}. 
Thus, proper journalistic rigor has come under threat through the online spread of fake news.

\citet{Wang2017} employed SVM (Support Vector Machine), LogReg (Logistic Regression), BiLSTM (Bidirectional Long Short-Term Memory), and CNN (Convolutional Neural Network), to detect the veracity of $\sim$13 K short statements. 
The preprocessing was done using Google News' pre-trained \textsc{Word2Vec} embeddings.
\citet{Conroy2015} present analysis methods based on linguistic and syntactic features for discovering fake news.

Many current approaches employ complex Deep Neural Network architectures, e.g., based on CNN (Convolutional Neural Network)~\citep{Kaliyar2020, Goldani2021, Saleh2021}, BiLSTM (Bidirectional Long Short-Term Memory)~\citep{Samantaray2022}, and others.
\citet{Ilie2021} used multiple deep neural networks to determine how models that use pre-trained and specific trained word embeddings perform in the task of fake news detection.
Further, some solutions use advanced document embeddings based on encoder architectures~\citep{Jwa2019}.
\citet{Kaliyar2021} propose FakeBERT, an extension of FNDNet that uses BERT instead of \textsc{GloVe} embeddings.
\citet{Kula2020} used a hybrid architecture for fake news detection that connects BERT with recurrent networks while
\citet{Mersinias2020} introduced CLDF, a new vectorization technique for extracting feature vectors.
The results for CLDF, FNDNet, and FakeBERT were obtained using the Kaggle dataset with $\sim$21~K news~articles.

Different ensemble models have also been used for this task, with good results~\citep{Mondal2022, Aslam2021}.
\citet{Mondal2022} used a voting-based ensemble method that relies on the voting of the collective majority.
The authors employ only non-deep learning models and TF-IDF as the vectorization technique.
\citet{Aslam2021} used an ensemble-based deep learning model that combines two architectures, i.e., Bi-LSTM-GRU-Dense and Dense.
\citet{Truica2022} propose MisRoBÆRTa, a BERT- and \textsc{RoBERTa}-based ensemble model for fake news detection.

\citet{Sedik2022} propose a deep learning approach that uses both sequential and recurrent layers. 
The sequential models employ stacked CNNs (i.e., CNN model) or concatenated CNN (i.e., C-CNN model), while the recurrent models use stacked CNN with LSTM and Dense layers (i.e., CNN-LSTM model)  or simple GRU with a Dense layer (i.e., GRU model).
The experimental results using the binary labeled Kaggle and Fake News Challenge dataset show that C-CNN and CNN-LSTM have the best performance, i.e., C-CNN obtains an accuracy of 99.90\% on the Kaggle dataset, and CNN-LSTM obtains an accuracy of 96\% on the Fake News Challenge dataset.

Several current solutions are based on linguistic and syntactic features, e.g., WELFake~\citep{Verma2021}, which uses word embedding over linguistic features. 
In other current directions, multimodal learning that integrates comments~\citep{Shu2019a}, images~\citep{Khattar2019}, and the social and network context has been used~\citep{Shu2019a,Zhang2020,Yang2019}.
\citet{Wang2020} propose a knowledge-driven Multimodal Graph Convolutional Network model for detecting fake news from textual and visual information.
This solution models posts from social media as graph data structures that combine textual and visual data with knowledge concepts.

\citet{Le2014Doc2Vec} propose Doc2Vec as an extension to Word2Vec.
Doc2Vec computes a feature vector for every document in the dataset, as opposed to Word2Vec, which computes every word in the dataset.
Several articles have discussed the use of Doc2Vec for fake news detection, but it is used only as a baseline combined with traditional Machine Learning solutions.
\citet{Cui2022} use as a baseline Doc2Vec with SVM and compare it with graph-based Deep Learning solutions.
\citet{Singh2020} presents several experiments on LIAR and Kaggle datasets using different vector space representations, i.e., one-hot encoding, TFIDF, Word2Vec, and Doc2Vec.
\citet{Truica2022b} propose a BiLSTM architecture with Sentence Transformer for the fake news detection challenge at CheckThatLab! 2022.
The proposed architecture uses BART for a monolingual fake news detection task and XML-RoBERTa for the multilingual task.
For the multilingual task, the model relies on transfer learning.
Thus, the BiLSTM XML-RoBERTa model is trained on English and tested on a German dataset. 
The proposed model managed to obtain an accuracy of 0.53 for the first task and an accuracy of 0.28 for the second task. 

\section{Methodology}\label{sec:methodology}

\begin{figure}[!htbp]
        \centering
        \includegraphics[width=1\textwidth]{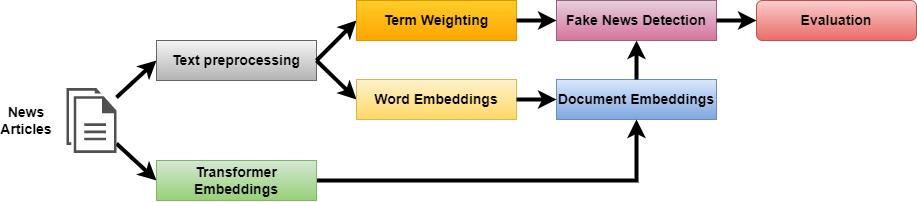}
    \caption{Proposed Pipeline.}
    \label{fig:workflowDiagram}
\end{figure}

Figure~\ref{fig:workflowDiagram} presents the pipeline of our proposed solution.
A labeled corpus of news articles is first preprocessed to extract the tokens.
Then, the tokens are transformed into a vector model using term weighting schemes (TFIDF) and word/transformer embeddings.
These vectors are used to create document embeddings.
We also use the raw corpus to create document embeddings using transformers.
The vectorized documents are then passed to the classification module.
Finally, the classification is evaluated using accuracy, precision, and recall.

\subsection{Text Preprocessing}

To prepare the text for vectorization, we use the following preprocessing steps to minimize the vocabulary and remove terms that bring no information gain~\citep{Truica2021b}: (1) removal of punctuation and stopwords, and (2) word lemmatization.
We chose to lemmatize the words to minimize the vocabulary and remove any language inflections.
We do not apply these preprocessing steps when using the transformer embeddings.

\subsection{Term Weighting}
To vectorize the preprocessed documents, we employ the TFIDF (Equation~\eqref{eq:tfidf}).
To compute this metric, we first need to compute the 
(1) term-frequency TF (Equation~\eqref{eq:tf}) and 
(2) the inverse document frequency (Equation~\eqref{eq:idf}).
For a set of $n$ documents $ D = \{ d_{i} \mid i \in \overline{1,n} \} $,
we extract the set of $m$ unique terms $V = \{ t_{j} \mid j \in \overline{1,m} \}$.
This set of unique terms is called a vocabulary.
For each term, we compute the raw frequency ($f_{t_{j},d_{i}}$), which counts the number of occurrences of a term $t_j$ in a document $d_i$.
The $f_{t_{j},d_{i}}$ does not store context and is biased towards longer documents. Thus, to remove the bias, we normalize the frequency with the length of the document ($\sum _{t'\in d_i}{f_{t',d_i}}$) and obtain TF~\citep{Paltoglou2010}.
Furthermore, to minimize the importance of common terms that bring no information value, the IDF (Equation~\eqref{eq:idf}) is used to reduce the TF weight by a factor that grows with the collection frequency $n_{j}$ of a term $t_{j}$, i.e., $n_{j}$ is the number of documents where there is at least one occurrence of term $t_{j}$.
Finally, to normalize TFIDF in the $[0, 1]$ range, we use the $\ell^{2}$-norm (Equation~\eqref{eq:l2norm}).
\begin{equation}
\label{eq:tfidf}
TFIDF(t_{j}, d_{i}, D) = \frac{TF(t_{j}, d_{i}) \cdot IDF(t_{j}, D)}{\ell^{2}(d_{i})}
\end{equation}
\begin{equation}
\label{eq:tf}
TF(t_{j}, d_{i}) = \frac{f_{t_{j},d_{i}}}{\sum _{t'\in d_{i}}{f_{t',d_{i}}}}
\end{equation}
\begin{equation}
\label{eq:idf}
IDF(t_{j}, D) = \log {\frac {n}{n_{j}}}
\end{equation}
\begin{equation}\label{eq:l2norm}
    \ell^{2}(d_{i}) = \sqrt{\sum_{t \in d_i}\left(TF(t, d_{i}) \cdot IDF(t, D)\right)^{2}}
\end{equation}

Using TFIDF, we construct a $n \times m$ document-term matrix $X = \{ x_{ij} \mid i = \overline{1,n} \wedge j = \overline{1,m} \}$ ($X \in \mathbb{R}^{n \times m}$) where rows correspond to documents and columns to terms.
The value $x_{ij} = TFIDF(t_{j}, d_{i}, D)$ represents the weight of term $t_{j}$ in document $d_{i}$.
Thus, each document $d_{i}$ is represented by a vector $\mathbf{x}_{i} = \{ x_{ij} \mid j \in \overline{1,m} \}$.
For ease of notation, we use $\mathbf{x}_{i}$ for denoting lines in $X$.

\subsection{Word Embeddings} 
Each word from the vocabulary is transformed into its vector representation.
This module employs \textsc{Word2Vec}~\citep{Mikolov2013a,Mikolov2013b}, \textsc{FastText}~\citep{Bojanowski2017}, and \textsc{GloVe}~\citep{Pennington2014}. 
For \textsc{Word2Vec} and \textsc{FastText}, we use both the CBOW (Continuous Bag-of-Words) and SG (Skip-gram) models.
By using these models, we obtain the embedding $WordEmb(t)$ for each term $t \in V$.

\subsubsection{\textsc{Word2Vec}}
The \textsc{Word2Vec}~\citep{Mikolov2013a,Mikolov2013b} embedding model is used to create vectorized representations of the words in a dataset within the same vector space.
This representation measures the distance between the corresponding vectors in this space to determine the context similarity.
For \textsc{Word2Vec}, there are two models for representing the words in this vector space: Continuous Bag-Of-Words (CBOW) or Skip-Gram.

\paragraph{CBOW Model}
The CBOW model attempts to predict a word using the context given by its surrounding words.
Each word $t_{i} \in V$ ($i \in \overline{1, m}$) is defined by two $d$-dimensional {(with $d \geq 2$ a natural number, i.e., $d \in \mathbb{N}$)} vectors depending of its function in training:
(1) $v_{t_{i}} \in \mathbb{R}^{d}$ is defined when $t_{i}$ is used as the center word, and
(2) $u_{t_{i}} \in \mathbb{R}^{d}$ is defined when $t_{i}$ is used as a context word.
The conditional probability of generating any center word $t_{c}$ given its surrounding context words $\mathcal{T}_{o} = \{ t_{1}, \ldots, t_{c-1}, t_{c+1}, \ldots, t_{s} \}$ within a context window of size $s$ can be modeled by a probability distribution $p(t_{c} \mid \mathcal{T}_{o})$ (Equation~\eqref{eq:w2w_cbow}) that considers the average of the context vectors $\overline{v}_{o}=\frac{1}{s}(v_{t_{1}} + \ldots + v_{t_{c-1}} + v_{t_{c+1}} + \ldots + v_{t_{s}})$.
\begin{equation}\label{eq:w2w_cbow}
  p(t_{c} \mid \mathcal{T}_{c}) = \frac{e^{u^{\top}_{c} \overline{v}_{o}}}{\sum_{i=1}^{m}e^{u^{\top}_{i}\overline{v}_{o}}}
\end{equation}

\paragraph{Skip-Gram Model}
The Skip-Gram model starts with the context word $t_{c}$ as input and tries to generate its context.
As in the CBOW case, the two $d$-dimensional vectors {($d \in \mathbb{N}$ and $d \geq 2$)}, i.e., $v_{t_{i}} \in \mathbb{R}^{d}$ and $u_{t_{i}} \in \mathbb{R}^{d}$, are defined for each word $t_{i} \in V$ ($i \in \overline{1, m}$).
The conditional probability of generating any context word $t_{o}$ given the center word $t_{c}$ can be modeled by a softmax operation (Equation~\eqref{eq:w2v_sg}).
\begin{equation}\label{eq:w2v_sg}
    p(t_{o} \mid t_{c}) =  \frac{e^{u^{\top}_{o} v_{c}}}{\sum_{i=1}^{m}e^{u^{\top}_{i}v_{c}}}
\end{equation}

\subsubsection{\textsc{FastText}}
\textsc{FastText}~\citep{Bojanowski2017} is an extension to \textsc{Word2Vec} and follows a similar approach to construct word embeddings~\citep{Mikolov2018}. 
The main difference between \textsc{FastText} and \textsc{Word2Vec} is that \textsc{FastText} does not consider the word as the basic unit, but rather considers a bag of character n-grams. 
Using such an approach, the accuracy is improved, and the training time is decreased when compared to \textsc{Word2Vec}.
As in the case of \textsc{Word2Vec}, \textsc{FastText} employs both CBOW and Skip-Gram models.

\subsubsection{\textsc{GloVe}}
\textsc{GloVe} (Global Vectors)~\citep{Pennington2014} is another model used for creating word embeddings.
To create the vector representation of words, \textsc{GloVe} uses the word co-occurrences matrix.
This matrix manages to encapsulate local and global corpus statistics regarding word--word co-occurrences.
Thus, \textsc{GloVe} for each word stores the frequency of its appearance in the same context as another word by employing a term co-occurrence matrix.
Using the ratio of co-occurrence probability, \textsc{GloVe} captures the relationship between words.
Furthermore, \textsc{GloVe} identifies word analogies and synonyms within the same contexts using this probability ratio.

\subsection{Transformers Embeddings} 
To create transformer embeddings, we use BERT~\citep{Devlin2019}, \textsc{RoBERTa}~\citep{Liu2019}, and BART~\citep{Lewis2020}.
By using these models, we obtain the word embedding by transformer $WordEmb(t)$ for each term $t \in V$.

\subsubsection{BERT}
BERT (bidirectional encoder representations from transformers)~\citep{Devlin2019} is a deep bidirectional transformer architecture used for natural language understanding.
Thus, in contrast to classic language models that treat textual data as unidirectional or bidirectional sequences of words, BERT learns contextual relations between the words by employing this deep bidirectional transformer architecture. 
Using the surrounding words of a given word, the model learns and creates a vector representation for each word that also encapsulates its context.
Thus, BERT reads the entire sequence of words at once using the transformer encoder to create contextual word embeddings.
By employing transfer learning, BERT can directly be used for various natural language processing operations, understanding, and generation.
Furthermore, it can be fine-tuned by using new datasets to adapt to specific tasks.
Experimental results on various tasks~\citep{Devlin2019} show that the language models built with BERT manage to improve language context detection more than the models that use static word embeddings, which only see textual data as sequences of words.

\subsubsection{\textsc{RoBERTa}}
\textsc{RoBERTa} (a Robustly optimized BERT pre-training Approach)~\citep{Liu2019} is a training optimizing method for BERT.
This model improves the language masking strategy of BERT by modifying the following key training aspects:
(1) more data are used for training,
(2) dynamic masking patterns instead of static masking patterns are employed,
(3) the next-sentence pre-training objective is removed and replaced with full sentences without NSP (Next Sentence Prediction),
(4) training is performed on longer sequences,
(5) the mini-batches are improved, and 
(6) the learning rates are improved.
Thus, all these modifications lead to improving \textsc{RoBERTa}'s downstream task performance and mitigate some of the shortcomings encountered by the significantly undertrained BERT model.

\subsubsection{BART}
BART (bidirectional and autoregressive transformer)~\citep{Lewis2020} is a transformer model that employs the standard transformer-based neural machine translation architecture, i.e., a generalized BERT architecture.
The pre-training process of BART uses an arbitrary noising function to corrupt the textual data within the dataset in order to make the transformer learn how to recreate the original text during training. 
During the pre-training of BART, two key techniques are used to improve the words' contextual representations.
Firstly, the order of original sentences is randomly shuffled. 
Secondly, using a novel in-filling scheme, a single mask token is used to replace the spans of text.
Experimental results~\citep{Lewis2020} show that a fine-tuned BART works better than BERT for both text generation and comprehension~tasks.

\subsection{Document Embeddings}
We create a vector for each document by averaging all the word or transformer embeddings for the words appearing in the document.
Thus, if we have $m_{i}$ terms in a document $d_{i}$, we obtain the document embedding (\textsc{DocEmb}) $x_i$ by summing all the embeddings $w(t)$ of the terms $t$ that are present in document $d_{i}$ as well as in the vocabulary $V$, and dividing the sum by $m_{i}$ (Equation~\eqref{eq:docemb}).
Each document embedding creates a context for words in a document and becomes an extension of the presented word embeddings.\vspace{-3pt}
\begin{equation}
\label{eq:docemb}
\mathbf{x}_i = \frac{\sum _{t\in d_{i}}{w(t)}}{m_{i}}
\end{equation}

Similarly to TFIDF, we construct a document-embedding matrix $X = \{ \mathbf{x}_{i} \mid i = \overline{1,n} \}$ where each row corresponds to the document embedding $\mathbf{x}_{i}$.
For this matrix, the columns are not associated to terms in the vocabulary $V$, and the number of columns is different from the total number of terms in $V$.
In this case, $m$ is the size of the embedding vector.
For ease of notation, we use $m$ as the number of columns, although it is different than the number of terms in the vocabulary, as in the case of the document-term matrix.
Thus, {$X \in \mathbb{R}^{n \times m}$} is a $n \times m$ matrix.

\subsection{Fake News Detection}
Classification is used to determine the veracity of a news article, {i.e, fake news detection}.
Given a set of documents $D$ represented by the matrix $X \in \mathbb{R}^{n \times m}$ (either the document-terms or the document embedding matrix), a set of classes $Y = \{ y_{1}, \ldots, y_{n} \}$ with values in a discrete domain $C = \{ c_{k} \mid k = \overline{1,\kappa} \}$ ($Y \subseteq C$) of size $\kappa$ (i.e., the number of classes is $\kappa$), and an implication $\mathbf{x}_{i} \rightarrow y_{i}$ ($ i \in \overline{1, n)}$, the objective of classification is to predict $\hat{y}_{i} = f(\mathbf{x}_{i})$ ($\hat{y}_{i} \in \hat{Y} \subseteq C$) that best approximates $y_{i}$.
{For the fake news detection task, we employ the following algorithms to construct models}: Naïve Bayes (NB), Gradient Boosted Trees (XGBTrees), Perceptron, Multi-Layer Perceptron (MLP), Long Short-Term Memory Network (LSTM), Bidirectional LSTM (BiLSTM), Gated Recurrent Units (GRU), and Bidirectional GRU (BiGRU).
For comparison, we use MisRoBÆRTa~\citep{Truica2022}.
In the original article that presents MisRoBÆRTa, the authors fine-tune both BART and \textsc{RoBERTa}.
In this work, we use the pre-trained BART (\textit{facebook/bart-large}) and \textsc{RoBERTa} (\textit{roberta-base}) from HuggingFace~\citep{Wolf2020}.

\subsubsection{Naïve Bayes}
The Naïve Bayes (NB) model is a probabilistic classification algorithm that computes the probability of $\mathbf{x}_{i}$ given a class $c_{k}$ (Equation~\eqref{eq:nb}), where $p(\mathbf{x}_{i})$ and $p(c_{k})$ are the probability of a document and a class, respectively, {and} $p(\mathbf{x}_{i} \mid c_{k})$ is the probability of class $c_{k}$ given $\mathbf{x}_{i}$.
Expending $\mathbf{x}_{i}$ by its components $\{ x_{i1}, \ldots, x_{im} \}$, we can rewrite Equation~\eqref{eq:nb} as Equation~\eqref{eq:nb_expanded}.
\begin{equation}
\label{eq:nb}
    p(c_{k} \mid \mathbf{x}_{i}) = \frac{p(c_{k}) p(\mathbf{x}_{i} \mid c_{k})}{p(\mathbf{x}_{i})}
\end{equation}
\begin{equation}
\label{eq:nb_expanded}
p(c_{k} \mid x_{i1}, \ldots, x_{im}) = \frac{p(c_{k})p(x_{i1}, \ldots, x_{im} \mid c_{k})}{p(\mathbf{x}_{i})}
\end{equation}

The denominator $p(\mathbf{x}_{i})$ is constant, while $p(x_{i1}, \ldots, x_{im}\mid c_{k})$ is equivalent to the joint probability $p(c_{k}, x_{i1}, \ldots, x_{im})$.
Furthermore, all the terms are conditionally independent given a class $c_{k}$.
Thus, $p(c_{k}, x_{i1}, \ldots, x_{im}) = \prod_{j=1}^{m}p(x_{ij} \mid c_{k})$.
Using these assumptions, the Naïve Bayes classifier tries to estimate the class $\hat{y}_{i}$ using Equation~\eqref{eq:nb_estimator}. 
\begin{equation}\label{eq:nb_estimator}
    \hat{y}_{i} = \argmax_{c_{k} \in C} p(c_{k})\prod_{j=1}^{m}p(x_{ij} \mid c_{k})
\end{equation}

There are various types of Naïve Bayes classifiers; the most common ones are Multinomial Naïve Bayes and Gaussian Naïve Bayes.

\paragraph{Multinomial Naïve Bayes}
Multinomial Naïve Bayes (MNB) models the distribution of words in a document by using a multinomial representation for the distribution of probabilities that a word appears for a certain class (Equation~\eqref{eq:mnb_pwc}).
The assumption for this model is that a document is handled as a sequence of words. Also, it is assumed that each word position is generated independently of every other~\citep{Rennie2003}.\vspace{-3pt}
\begin{equation}\label{eq:mnb_pwc}
    p(\mathbf{x}_{i} \mid c_{k}) = \frac{(\sum_{j=1}^{m}x_{ij})!}{\prod_{j=1}^{m}x_{ij}!}\prod_{j=1}^{m}p(x_{ij}\mid c_{k})^{x_{ij}}
\end{equation}

Equation~\eqref{eq:mnb} presents the Multinomial Naïve Bayes classification model.
\begin{equation}\label{eq:mnb}
\begin{split}
    \hat{y}_{i} = \argmax_{c_{k} \in C} p(c_{k})  \frac{(\sum_{j=1}^{m}x_{ij})!}{\prod_{j=1}^{m}x_{ij}!}\prod_{j=1}^{m}p(x_{ij}\mid c_{k})^{x_{ij}}
\end{split}
\end{equation}

\paragraph{Gaussian Naïve Bayes} 
The Gaussian Naïve Bayes (GNB) model is used when dealing with continuous data. 
The model is based on the assumption that continuous values correlated with each class are distributed according to a Gaussian distribution.
Thus, given column $j \in \overline{1,m}$ from $X$ and a class $c_{k}$, we employ the following steps:
\begin{itemize}
    \item Segment the data by class $c_{k}$.
    \item Compute the associated means $\mu_{j}$ and variances $\sigma_{j}$ of dimension $j$ using the values $x_{ij}$ ($i \in \overline{1, n}$), only for the lines $\mathbf{x}_{i} \in X$ labeled with class $c_{k}$.
    \item Compute the probability $p(x_{ij} \mid c_{k})$ (Equation~\eqref{eq:gnb1}).
\end{itemize}
\begin{equation}\label{eq:gnb1}
    p(x_{ij} \mid c_{k}) = \frac{1}{\sqrt{2\pi\sigma_{j}^2}} e^{-\frac{x_{ij} - \mu_{j}^2}{2 \sigma_{j}^2}}
\end{equation}

Equation~\eqref{eq:gnb} presents the Gaussian Naïve Bayes classifier.
\begin{equation}\label{eq:gnb}
    \hat{y}_{i} = \argmax_{c_{k} \in C} p(c_{k}) \prod_{j=1}^{m} \frac{1}{\sqrt{2\pi\sigma_{j}^2}} e^{-\frac{x_{ij} - \mu_{j}^2}{2 \sigma_{j}^2}}
\end{equation}

\subsubsection{Gradient Boosted Trees}
Gradient boosting is an ensemble method that uses multiple weak predictions learners.
In the case of Gradient Boosted Trees, the weak learners are Decision Trees.
Similar to other classification methods, the method tries to predict $\hat{y}_{i} = f(\mathbf{x}_{i}) = \mathbf{w} \cdot \mathbf{x}_{i} + b$ that best approximates the true class $y_{i}$ of $\mathbf{x}_{i}$ by minimizing an objective function $L(\hat{Y}_i, Y_{i})$ that represents the training loss, e.g., the mean score $L(y_{i}, \hat{y}_i) = \frac{1}{n}\sum_{i=1}^{n}(\hat{y}_i - y_{i})^{2} = \frac{1}{n}\sum_{i=1}^{n}l(\hat{y}_i,y_{i})$.
As the model builds $T$ weak learners $f^{(t)}(\mathbf{x}_{i})$ ($t \in \overline{1,T}$), at each stage $t \in T$ the model tries to determine $\hat{y}^{(t)}_{i} = \hat{y}^{(t-1)}_{i} + f^{(t)}(\mathbf{x}_{i})$ using the previously estimated value $\hat{y}^{(t-1)}_{i}$ and the function $f^{(t)}(\mathbf{x}_{i})$ determined by the current weak learner that best fits the residuals, i.e., $f^{(t)}(\mathbf{x}_{i}) = y_{i} - \hat{y}^{(t)}_{i}$.
As the objective is to minimize training loss and to obtain the specific objective at step $t$,we can take the Taylor expansion of the loss function up to the second order  for each learner and remove all the constants to obtain $L^{(t)}(\hat{Y}^{(t)}_i, Y_{i})$ (Equation~\eqref{eq:obj_gbt}).
\begin{equation}\label{eq:obj_gbt}
\begin{split}
    L^{(t)}(\hat{Y}^{(t)}_i, Y_{i}) & = \sum_{i=1}^{n} [ g_{i}f^{(t)}(\mathbf{x}_{i}) +  \frac{1}{2} h_{i} (f^{(t)}(\mathbf{x}_{i}))^{2}] \\
    g_{i} & = \partial_{\hat{y}^{(t-1)}_{i}} l (y_{i}, \hat{y}^{(t-1)}_{i}) \\
    h_{i} & = \partial_{\hat{y}^{(t-1)}_{i}}^{2} l (y_{i}, \hat{y}^{(t-1)}_{i})
\end{split}
\end{equation}

\subsubsection{Perceptron} 
The Perceptron model (Equation~\ref{eq:perceptron}) is a simple non-linear processing unit that tries to predict the label $\hat{y}_i$ for a given input $x_{i}$ by adjusting a weight vector $\mathbf{w} \in \mathbb{R}^{m}$ using the sigmoid activation $\delta_{s}(z)=\frac{1}{1+e^{-z}} \in [0, 1]$.
The objective for a good prediction is to minimize the average cross-entropy loss function between the set of prediction $\hat{Y}$ and the set of true labels $Y$ (Equation~\ref{eq:loss_fn}).
\begin{equation}\label{eq:perceptron}
    \hat{y}_i=\delta_{s}(\mathbf{w} \cdot \mathbf{x}_{i} + b)
\end{equation}
\begin{equation}\label{eq:loss_fn}
    L(\hat{Y}, Y) = -\frac{1}{n}\sum_{i=1}^{n}(y_i\log\hat{y}_i + (1 - y_i) \log(1 - \hat{y}_i))
\end{equation}

\subsubsection{Multi-Layer Perceptron} 
The Multi-Layer Perceptron (MLP) model is a Deep Learning architecture that stacks multiple layers $j \in \overline{1,l}$ of fully-connected Perceptron units.
The MLP architecture can be divided into three components: 
(1) the input $i$ layer ($j=1$),
(2) the hidden layers $j \in \overline{2,l-1}$, and
(3) the output layer $o = \hat{y}$ ($j=l$).
Each node in layer $j$ connects to every node in the following layer $j+1$ with a certain weight $W_{j}$.
Because the connections between the layers are directed from the input $i$ to the output $o$ by passing information through the hidden layers $h_{j}$, the MLP model is a feed-forward architecture.
Equation~\eqref{eq:mlp} presents the MLP classification model at a given iteration $t$.
\begin{equation}\label{eq:mlp}
    \begin{split}
    i^{(t)}       & = \delta_{s}(W_{1} \cdot \mathbf{x}^{(t)}_{i} + b_{1}) \\
    h^{(t)}_{j}   & = \delta_{s}(W_{j} \cdot h^{(t)}_{j-1} + b_{j}) \\
    o^{(t)} & = \delta_{s}(W_{l} \cdot h^{(t)}_{l-1} + b_{l})
    \end{split}
\end{equation}

\subsubsection{Long Short-Term Memory}
Long Short-Term Memory (LSTM)~\citep{Hochreiter1997} is a Recurrent Artificial Neural Network that uses two state components for classification.
The first component, represented by a hidden state, is the short-term memory that learns the short-term dependencies between the previous and current states.
The second component, represented by an internal cell state, is the long-term memory which stores long-term dependencies between the previous and current states.
The model uses three gates to preserve the long-term memory within the state:
(1) input gate ($i \in \mathbb{R}^{n}$),
(2) forget gate ($f \in \mathbb{R}^{n}$), and
(3) output gate ($o \in \mathbb{R}^{n}$).
Equation~\eqref{eq:lstm} presents the compact forms for the state updates of the LSTM unit for a given iteration $t$, where:
\begin{itemize}
 \item $\mathbf{x}^{(t)}_{i} \in \mathbb{R}^m$ is the input vector of dimension $m$ at step $t$, with $\mathbf{x}^{(0)}_{i} = \mathbf{x}_{i} \in X$; 
 \item $h^{(t)} \in \mathbb{R}^n$ is the hidden state vector as well as the unit's output vector of dimension $n$, where the initial value is $h^{(0)}=0$;
 \item $\Tilde{c}^{(t)} \in \mathbb{R}^n$ is the input activation vector;
 \item $c^{(t)} \in \mathbb{R}^n$ is the cell state vector, with the initial value $c^{(0)}=0$;
 \item $W_i, W_f, W_o, W_c \in \mathbb{R}^{n \times m}$ are the weight matrices corresponding to the current input of the input gate, output gate, forget gate, and the cell state;
 \item $V_i, V_f, V_o, V_c \in \mathbb{R}^{n \times n}$ are the weight matrices corresponding to the hidden output of the previous state for the current input of the input gate, output gate, forget gate, and the cell state;
 \item $b_i, b_f, b_o, b_c \in \mathbb{R}^{n}$ are the bias vectors corresponding to the current input of the input gate, output gate, forget gate, and the cell state;
 \item $\delta_h(z)=\frac{e^{z} - e^{-z}}{e^{z} + e^{-z}} \in [-1, 1]$ is the hyperbolic tangent activation function;
 \item $\odot$ is the Hadamard Product, i.e., element wise product.
\end{itemize}
\begin{equation}\label{eq:lstm}
\begin{split}
i^{(t)} & = \delta_{s}(W_{i} \mathbf{x}^{(t)}_{i} + V_{i} h^{(t-1)} + b_{i}) \\
f^{(t)} & = \delta_{s}(W_{f} \mathbf{x}^{(t)}_{i} + V_{f} h^{(t-1)} + b_{f})\\
o^{(t)} & = \delta_{s}(W_{o} \mathbf{x}^{(t)}_{i} + V_{o} h^{(t-1)} + b_{o})\\
\Tilde{c}^{(t)} & = \delta_{h}{(W_{c} \mathbf{x}^{(t)}_{i} + V_{c} h^{(t-1)} + b_c)} \\
c^{(t)} & = i^{(t)} \odot \Tilde{c}^{(t)} + f^{(t)} \odot c^{(t-1)}\\
h^{(t)} & = o^{(t)} \odot \delta_{h}(c^{(t)})
\end{split}
\end{equation}

We chose LSTM because it manages to avoid the vanishing and the exploding gradient issues by regulating the way the recurrent weights are learned.

\subsubsection{Bidirectional LSTM}
As the LSTM model processes sequence data, it is able to capture \textit{past} information.
To take into consideration \textit{future} information as well, we use the Bidirectional LSTM (BiLSTM).
The BiLSTM encapsulates \textit{past} and \textit{future} information through the use of two hidden states (Equation~\eqref{eq:bilstm}), where
(1) $\overrightarrow{h}^{(t)}$ processes the input in a forward manner using the past information provided by the forward LSTM ($\overrightarrow{LSTM}_F$), and 
(2) $\overleftarrow{h}^{(t)}$ processes the input in a backward manner using the future information provided by the backward LSTM ($\overleftarrow{LSTM}_B$).
\begin{equation}\label{eq:bilstm}
\begin{split}
    \overrightarrow{h}^{(t)} & = \overrightarrow{LSTM}_{F}(\mathbf{x}^{(t)}_{i}) \\
    \overleftarrow{h}^{(t)}  & = \overleftarrow{LSTM}_{B}(\mathbf{x}^{(t)}_{i})
\end{split}
\end{equation}

At every time-step, the hidden states, i.e., $\overrightarrow{h}^{(t)}$ and $\overleftarrow{h}^{(t)}$, are concatenated into one hidden state $h'^{(t)}$ (Equation~\eqref{eq:concat}).
This approach enables the encoding of information from both past and future contexts in the hidden state $h'^{(t)}$.
\begin{equation}\label{eq:concat}
    h'^{(t)} = [\overrightarrow{h}^{(t)} \mid\mid \overleftarrow{h}^{(t)}]
\end{equation}

\subsubsection{Gated Recurrent Unit}
The Gated Recurrent Unit (GRU)~\citep{Cho2014} is a Recurrent Artificial Neural Network that simplifies the LSTM unit and improves performance considerably.
Instead of three gates as in the case of LSTM, the GRU has only two gating mechanisms.
The first gating mechanism is the update gate ($u \in \mathbb{R}^{n}$).
This gate encodes both the forget gate and the input gate that are present in the LSTM cell.
The second gating mechanism is the reset gate ($r \in \mathbb{R}^{n}$).
This gate determines the percentage of information from the previous hidden state that contributes to the candidate state of the new step~\citep{Hewamalage2021}
Furthermore, the GRU uses the hidden state as the only state component.
Equation~\eqref{eq:gru} presents the compact forms for the state updates of the GRU unit at a given iteration step $t$, where:
\begin{itemize}
    \item $\mathbf{x}^{(t)}_{i} \in \mathbb{R}^{m}$ is the input vector of dimension $m$ at step $t$, with $\mathbf{x}^{(0)}_{i} = \mathbf{x}_{i} \in X$;
    \item $i^{(t)} \in \mathbb{R}^{n}$ is the input and output of the cell at step $t$;
    \item $\Tilde{h}^{(t)} \in \mathbb{R}^n$ is the candidate hidden state with a cell dimension of $n$;
    \item $h^{(t)} \in \mathbb{R}^n$ is the current hidden state with a cell dimension of $n$;
    \item $W_u, W_r, W_h \in \mathbb{R}^{n \times m}$ are the weight matrices corresponding to the current input of the update gate, reset gate, and the hidden state;
    \item $V_u, V_r, V_h \in \mathbb{R}^{n \times m}$ are the weight matrices corresponding to the hidden output of the previous state for the current input of the update gate, reset gate, and the hidden state;
    \item $b_u, b_r, b_h \in \mathbb{R}^{n}$ are the bias vectors corresponding to the current input of the update gate, reset gate, and the hidden state;
    \item $\odot$ is the Hadamard Product.
\end{itemize}
\begin{equation}\label{eq:gru}
\begin{split}
u^{(t)} & = \delta_{s}(W_{u} \mathbf{x}^{(t)}_{i} + V_{u} h^{(t-1)} + b_{u}) \\
r^{(t)} & = \delta_{s}(W_{r} \mathbf{x}^{(t)}_{i} + V_{r} h^{(t-1)} + b_{r})\\
\Tilde{h}^{(t)} & = \delta_{h}{(W_{h} \mathbf{x}^{(t)}_{i} + V_{h} h^{(t-1)} + b_{h})} \\
h^{(t)} & = u^{(t)} \odot \Tilde{h}^{(t)} + (1 - u^{(t)}) \odot h^{(t-1)}
\end{split}
\end{equation}

\subsubsection{Bidirectional GRU}
Similar to the BiLSTM, the Bidirectional GRU (BiGRU) considers both \textit{past} and \textit{future} information by employing a forward and backward GRU, i.e., $\overrightarrow{GRU}_F$ and $\overleftarrow{GRU}_B$, respectively. 
The $\overrightarrow{GRU}_F$ and $\overleftarrow{GRU}_B$ are associated to two hidden states (Equation~\eqref{eq:bigru}): 
(1) $\overrightarrow{h}^{(t)}$ which processes the input in a forward manner using $\overrightarrow{GRU}_F$, and 
(2) $\overleftarrow{h}^{(t)}$ which processes the input in a backwards manner using $\overleftarrow{GRU}_B$.
As for BiLSTM, the hidden states $\overrightarrow{h}^{(t)}$ and $\overleftarrow{h}^{(t)}$ are concatenated at every time-step to encode the information from both past and future contexts into one hidden state $h'^{(t)} =  [\overrightarrow{h}^{(t)} \mid\mid \overleftarrow{h}^{(t)}]$.
\begin{equation}\label{eq:bigru}
\begin{split}
    \overrightarrow{h}^{(t)} & = \overrightarrow{GRU}_{F}(\mathbf{x}_{i}) \\
    \overleftarrow{h}^{(t)}  & = \overleftarrow{GRU}_{B}(\mathbf{x}_{i})
\end{split}
\end{equation}

\subsection{Evaluation Module}

We use accuracy, precision, and recall~\citep{Sokolova2009} to evaluate the models.
For binary classification with the classes positive and negative, the following information is used to construct a confusion matrix that is afterward used to compute the evaluation metrics:
\begin{itemize}
    \item $tp$ (True Positive) is the number of positive observations that are correctly classified;
    \item $fn$ (False Negative) is the number of positive observations that are incorrectly classified as negative;
    \item $fp$ (False Positive) is the number of false observations that are incorrectly classified as positive;
    \item $tn$ (True Negative) is the number of false observations that are correctly classified.
\end{itemize}

Accuracy (Equation~\eqref{eq:acc}) measures the overall effectiveness of a classifier.
Precision (Equation~\eqref{eq:prec}) measures the class agreement of the data labels within the positive labels.
Recall (Equation~\eqref{eq:rec}) measures the effectiveness of a classifier in identifying positive labels.
\begin{equation}\label{eq:acc}
    A=\frac{tp + tn}{tp + tn + fp + fn}
\end{equation}
\begin{equation}\label{eq:prec}
    P=\frac{tp}{tp + fp}
\end{equation}
\begin{equation}\label{eq:rec}
    R=\frac{tp}{tp + fn}
\end{equation}

\section{Experimental Results}\label{sec:results}

In this section, we present the experimental results obtained using our methodology.
Firstly, we introduce a human-verified sample from the Fake News Corpus~\citep{FakeNewsCorpus} and present the results of the exploratory data analysis performed on it.
Secondly, we present the experimental setup for our experiments as well as the hyperparameters and implementation packages for the models.
Thirdly, we present the experimental results using the different sentence embeddings and classification methods on the Fake News Corpus sample.
Lastly, we show the generalization of our observation by performing additional experiments on 5 additional datasets: LIAR multiclass~\citep{Wang2017}, LIAR binary~\citep{Upadhayay2020}, Kaggle~\citep{Kaliyar2020,Kaliyar2021}, Buzz Feed News~\citep{Horne2017}, and TSHP-17~\citep{Rashkin2017,BarronCedeno2019a}.

\subsection{Dataset Details} 

For the experiments, we used a set of {20K English language news articles (10K reliable and 10K fake) selected from the Fake News Corpus~\citep{FakeNewsCorpus} as it is widely used in current research~\citep{Kurasinski2020,Noerregaard2019,Kwak2020,Ilie2021,Truica2022}.
Some of the labels might not be correct because the original dataset is not manually annotated.
However, this shortcoming should not pose a practical issue for classification models, as ML/DL models generalize better when noise is added~\citep{Reed1999}.
Instead, this should help the models to better generalize and remove overfitting.
Additionally, we made sure that  URL for the selected article point to the correct article by matching the titles and authors.

Before performing the experiments, we verified the label correctness for the sampled news articles using computer science students as annotators.
In total, there were 40 student annotators to annotate 25K articles (12.5K reliable and 12.5K fake). 
We sampled more articles to mitigate any inconsistencies between two annotators as well as between the final annotation and the original label.
For their annotation work, the students obtained credits for different courses.

Before annotating the articles, the students received an instruction list that explains the annotation task.
The annotation task included the following steps:
\begin{itemize}
    \item[(1)] Verify that the title matches the title from the URL;
    \item[(2)] Verify that the content matches the content from the URL;
    \item[(3)] Verify that the authors match the authors from the URL;
    \item[(4)] Verify that the source matches the source from the URL;
    \item[(5)] Verify if the information is false or reliable;
\end{itemize}

Each article was manually verified by two annotators.
If there was no consensus between the two, a third annotator was used to break the tie.
In 99\% of the cases, there was no requirement for adding a third annotator.
In the experiments, we removed all the articles where no consensus was found as well as where a difference between the human annotation and the original label was found.
In the end, we scaled down the sample to $20K$ news articles.

Table~\ref{tab:classes} presents the corpus statistics and information before and after preprocessing.
We observe that, although there is a small imbalance in the number of tokens between the classes, this imbalance is small enough not to add bias to the classification task.
We also extracted the top 10 unigrams and the top topic  using the NMF algorithm for topic modeling~\citep{Arora2012}.
We used the \textit{NLTK}~\citep{Bird2009} for extracting unigrams and \textit{scikit-learn}~\citep{Pedregosa2011} for NMF.
We computed the average similarity with \textit{PolyFuzz}~\citep{PolyFuzz} by employing the pre-trained \textsc{FastText} embedding on news articles (\textbf{sim(FT)}) and the base-case BERT  (\textbf{sim(BERT)}).
Analyzing both similarities, we conclude that the documents discuss the same topics (Table~\ref{tab:classes}).
For the neural networks (i.e., Perceptron, MLP, LSTM, BiLSTM, GRU, BiGRU), we use one-hot encoders to represent the labels.

\begin{table*}[!htb]
\centering
\caption{Dataset description and statistics and information}
\label{tab:classes}
\resizebox{1\textwidth}{!}{
\begin{tabular}{|l|l|l|r|r|r|r|r|r|}
\hline
\multicolumn{3}{|l|}{\textbf{Statistics before preprocessing}}                                                                                              & \multicolumn{4}{c|}{\textbf{\#Tokens per document}}                                                                                               & \multicolumn{2}{c|}{\textbf{\#Tokens per Class}}                                \\ \hline
\textbf{Class}  & \textbf{Encoding}                                                                    & \textbf{Description}                             & \multicolumn{1}{c|}{\textbf{Mean}} & \multicolumn{1}{c|}{\textbf{Min}} & \multicolumn{1}{c|}{\textbf{Max}} & \multicolumn{1}{c|}{\textbf{StdDev}} & \multicolumn{1}{c|}{\textbf{Unique}}  & \multicolumn{1}{c|}{\textbf{All}}       \\ \hline
\textbf{Fake News} & 1                                                                 & Fabricated or distorted information & 517.83                             & 6                                 & 8\,812                            & 883.35                               & 119\,283                              & 5\,178\,300                             \\ \hline
\textbf{Reliable} & 0                                                                  & Reliable information                & 575.66                             & 7                                 & 10\,541                           & 602.16                               & 82\,203                               & 5\,756\,643                             \\ \hline
\multicolumn{3}{|l|}{\textbf{Entire dataset statistics}}                                                                                                    & 546.75                             & 6                                 & 10\,541                           & 618.63                               & 159\,113                              & 10\,934\,943                            \\ \hline  \hline
\multicolumn{7}{|l|}{\textbf{Textual information after preprocessing}}                                                                                                                                                                                                                                          & \multicolumn{1}{c|}{\textbf{sim(FT)}} & \multicolumn{1}{c|}{\textbf{sim(BERT)}} \\ \hline
\multirow{2}{*}{\textbf{\begin{tabular}[c]{@{}l@{}}Top-10\\ Unigrams\end{tabular}}} & \textbf{Fake News} & \multicolumn{5}{l|}{people time government world year story market American God day}                                                                                                                 & \multirow{2}{*}{\textbf{0.83}}        & \multirow{2}{*}{\textbf{0.93}}          \\ \cline{2-7}
                                                                                    & \textbf{Reliable}  & \multicolumn{5}{l|}{people God Christian government American time world war America political}                                                                                                       &                                       &                                         \\ \hline
\multirow{2}{*}{\textbf{\begin{tabular}[c]{@{}l@{}}Top-1\\ Topic\end{tabular}}}     & \textbf{Fake News} & \multicolumn{5}{l|}{people Trump year day government time state world market war}                                                                                                                    & \multirow{2}{*}{\textbf{0.84}}        & \multirow{2}{*}{\textbf{0.94}}          \\ \cline{2-7}
                                                                                    & \textbf{Reliable}  & \multicolumn{5}{l|}{church Trump people God president war state year Bush government}                                                                                                                &                                       &                                         \\ \hline
\end{tabular}
}
\end{table*}

\subsection{Experimental Setup}\label{subsec:experimental_setup}
In our experiments, we analyzed how well we can predict if an article is fake or reliable using multiple vectorizations:
(1) the TFIDF vector space model, and
(2) eight document embeddings (\textsc{DocEmb}) constructed using five word embeddings (\textsc{Word2Vec} CBOW, \textsc{Word2Vec} SG, \textsc{FastText} CBOW, \textsc{FastText} SG, and \textsc{GloVe}), and three transformers embeddings (BERT, \textsc{RoBERTa}, and BART).
We trained our own word embeddings and TFIDF vectorizer.
We trained each word embedding for 100 epochs using a window size of 10, a learning rate of 0.05, and a size of 128.
For TFIDF, we ignored words that appeared in less than 4 documents and kept the top {5K} 
relevant features.
We used \textit{SpaCy} for preprocessing, \textit{scikit-learn} for implementing TFIDF, \textit{gensim}~\citep{Rehurek2010} for the \textsc{Word2Vec} and \textsc{FastText} models, and the \textit{python-glove}~\citep{PyGloVe}
package for \textsc{GloVe}. 
For the transformer embeddings, we used the pre-trained uncased large versions from \textit{HuggingFace}~\citep{Wolf2020} together with \textit{SimpleTransformers}~\citep{Rajapakse2021} and \textit{SentenceTransformers}~\citep{Reimers2019}. 

For the experiments, we used  the following algorithms for classification: Naïve Bayes (NB), Perceptron, Multi-layer Perceptron (MLP), LSTM, Bidirectional LSTM (BiLSTM), GRU, and Bidirectional GRU (BiGRU).
We used \textit{scikit-learn} for implementing NB.
We applied the Multinomial NB for the TFIDF experiments as documents can get sparse, and the Gaussian NB for the transformer and word embeddings experiments as the embeddings can have negative values.
For the Gradient Boosted Trees, we used the XGBoost Python library~\citep{Chen2016}.

The neural-based fake news detection module used Multi-layer Perceptron, LSTM, Bidirectional LSTM, GRU, and Bidirectional GRU.
Each layer consists of 100 cells.
The LSTM was configured as in~\citep{Hochreiter1997}, while the GRU was configured as presented in~\citep{Cho2014}.
A Dense layer with 2 units and a sigmoid function as activation was used as the output.

For the LSTM, BiLSTM, GRU, and BiGRU models, we used the ADAM optimizer and a 64-batch size.
All the neural network models were trained for 100 epochs with an Early-Stopping mechanism to mitigate overfitting.
We employed \textit{Keras} for implementing the neural models.
For comparison, we used the free implementation of MisRoBÆRTa~\citep{Truica2022}, made available by the authors on GitHub.

The code will be made available on GitHub upon the acceptance of this work.

\subsection{Fake News Detection}

For the experiments, we used an NVIDIA\textsuperscript{\textregistered} DGX Station\textsuperscript{\texttrademark}.
Table~\ref{tab:results_fnc_full} presents the results for the fake news detection task.
We tested the models in 10 rounds and used a 70\%--10\%--20\% train--validation--test split ratio.
Each split shuffles the dataset and extracts a stratified sample initialized with different random seeds.
We report the average and standard deviation for each metric.

\begin{table*}[!htb]
\centering
\caption{Fake news detection results on the sample extracted from the Fake News Corpus}
\label{tab:results_fnc_full}
\resizebox{1\textwidth}{!}{%
\begin{tabular}{l|lll|lll|}
\cline{2-7}
 & \multicolumn{3}{c|}{\textbf{Naïve Bayes}} & \multicolumn{3}{c|}{\textbf{Gradient   Boosted Trees}} \\ \hline
\multicolumn{1}{|l|}{\textbf{Vectorization}} & \multicolumn{1}{c|}{\textbf{Accuracy}} & \multicolumn{1}{c|}{\textbf{Precision}} & \multicolumn{1}{c|}{\textbf{Recall}} & \multicolumn{1}{c|}{\textbf{Accuracy}} & \multicolumn{1}{c|}{\textbf{Precision}} & \multicolumn{1}{c|}{\textbf{Recall}} \\ \hline
\multicolumn{1}{|l|}{\textbf{TFIDF}} & \multicolumn{1}{l|}{92.69 $\pm$ 0.25} & \multicolumn{1}{l|}{91.72 $\pm$ 0.39} & 93.86 $\pm$ 0.41 & \multicolumn{1}{l|}{98.76 $\pm$ 0.21} & \multicolumn{1}{l|}{99.79 $\pm$ 0.08} & 97.72 $\pm$ 0.43 \\ \hline
\multicolumn{1}{|l|}{\textbf{\textsc{DocEmb} \textsc{Word2Vec} CBOW}} & \multicolumn{1}{l|}{66.29 $\pm$ 0.53} & \multicolumn{1}{l|}{60.78 $\pm$ 0.42} & 91.85 $\pm$ 0.27 & \multicolumn{1}{l|}{95.67 $\pm$ 0.18} & \multicolumn{1}{l|}{96.17 $\pm$ 0.37} & 95.13 $\pm$ 0.34 \\ \hline
\multicolumn{1}{|l|}{\textbf{\textsc{DocEmb} \textsc{Word2Vec} SG}} & \multicolumn{1}{l|}{53.10 $\pm$ 0.37} & \multicolumn{1}{l|}{51.74 $\pm$ 0.20} & 92.46 $\pm$ 0.77 & \multicolumn{1}{l|}{97.10 $\pm$ 0.21} & \multicolumn{1}{l|}{97.61 $\pm$ 0.31} & 96.56 $\pm$ 0.16 \\ \hline
\multicolumn{1}{|l|}{\textbf{\textsc{DocEmb} \textsc{FastText} CBOW}} & \multicolumn{1}{l|}{56.13 $\pm$ 0.53} & \multicolumn{1}{l|}{53.59 $\pm$ 0.33} & 91.45 $\pm$ 0.58 & \multicolumn{1}{l|}{94.90 $\pm$ 0.24} & \multicolumn{1}{l|}{95.49 $\pm$ 0.39} & 94.24 $\pm$ 0.46 \\ \hline
\multicolumn{1}{|l|}{\textbf{\textsc{DocEmb} \textsc{FastText} SG}} & \multicolumn{1}{l|}{54.00 $\pm$ 0.69} & \multicolumn{1}{l|}{52.23 $\pm$ 0.38} & 93.66 $\pm$ 0.98 & \multicolumn{1}{l|}{97.05 $\pm$ 0.26} & \multicolumn{1}{l|}{97.45 $\pm$ 0.32} & 96.64 $\pm$ 0.49 \\ \hline
\multicolumn{1}{|l|}{\textbf{\textsc{DocEmb} \textsc{GloVe}}} & \multicolumn{1}{l|}{53.43 $\pm$ 0.42} & \multicolumn{1}{l|}{51.98 $\pm$ 0.24} & 89.94 $\pm$ 0.78 & \multicolumn{1}{l|}{96.02 $\pm$ 0.30} & \multicolumn{1}{l|}{96.73 $\pm$ 0.35} & 95.26 $\pm$ 0.43 \\ \hline
\multicolumn{1}{|l|}{\textbf{\textsc{DocEmb} BERT}} & \multicolumn{1}{l|}{80.90 $\pm$ 0.64} & \multicolumn{1}{l|}{74.94 $\pm$ 0.67} & 92.87 $\pm$ 0.58 & \multicolumn{1}{l|}{97.43 $\pm$ 0.21} & \multicolumn{1}{l|}{97.75 $\pm$ 0.29} & 97.10 $\pm$ 0.30 \\ \hline
\multicolumn{1}{|l|}{\textbf{\textsc{DocEmb} \textsc{RoBERTa}}} & \multicolumn{1}{l|}{91.98 $\pm$ 0.31} & \multicolumn{1}{l|}{94.05 $\pm$ 0.44} & 89.63 $\pm$ 0.60 & \multicolumn{1}{l|}{97.38 $\pm$ 0.22} & \multicolumn{1}{l|}{98.72 $\pm$ 0.31} & 96.02 $\pm$ 0.42 \\ \hline
\multicolumn{1}{|l|}{\textbf{\textsc{DocEmb} BART}} & \multicolumn{1}{l|}{89.13 $\pm$ 0.32} & \multicolumn{1}{l|}{83.71 $\pm$ 0.46} & 97.19 $\pm$ 0.37 & \multicolumn{1}{l|}{98.26 $\pm$ 0.19} & \multicolumn{1}{l|}{98.18 $\pm$ 0.31} & 98.35 $\pm$ 0.21 \\ \hline
\textbf{} & \multicolumn{3}{c|}{\textbf{Perceptron}} & \multicolumn{3}{c|}{\textbf{Multi Layer Perceptron}} \\ \hline
\multicolumn{1}{|l|}{\textbf{Vectorization}} & \multicolumn{1}{c|}{\textbf{Accuracy}} & \multicolumn{1}{c|}{\textbf{Precision}} & \multicolumn{1}{c|}{\textbf{Recall}} & \multicolumn{1}{c|}{\textbf{Accuracy}} & \multicolumn{1}{c|}{\textbf{Precision}} & \multicolumn{1}{c|}{\textbf{Recall}} \\ \hline
\multicolumn{1}{|l|}{\textbf{TFIDF}} & \multicolumn{1}{l|}{95.79 $\pm$ 0.33} & \multicolumn{1}{l|}{96.55 $\pm$ 0.46} & 94.98 $\pm$ 0.42 & \multicolumn{1}{l|}{98.04 $\pm$ 0.19} & \multicolumn{1}{l|}{98.37 $\pm$ 0.28} & 97.70 $\pm$ 0.33 \\ \hline
\multicolumn{1}{|l|}{\textbf{\textsc{DocEmb} \textsc{Word2Vec} CBOW}} & \multicolumn{1}{l|}{93.61 $\pm$ 0.27} & \multicolumn{1}{l|}{94.22 $\pm$ 0.53} & 92.93 $\pm$ 0.49 & \multicolumn{1}{l|}{94.96 $\pm$ 0.31} & \multicolumn{1}{l|}{95.26 $\pm$ 0.36} & 94.63 $\pm$ 0.64 \\ \hline
\multicolumn{1}{|l|}{\textbf{\textsc{DocEmb} \textsc{Word2Vec} SG}} & \multicolumn{1}{l|}{92.04 $\pm$ 0.34} & \multicolumn{1}{l|}{94.65 $\pm$ 0.49} & 89.12 $\pm$ 0.91 & \multicolumn{1}{l|}{95.88 $\pm$ 0.30} & \multicolumn{1}{l|}{96.34 $\pm$ 0.65} & 95.40 $\pm$ 1.09 \\ \hline
\multicolumn{1}{|l|}{\textbf{\textsc{DocEmb} \textsc{FastText} CBOW}} & \multicolumn{1}{l|}{93.46 $\pm$ 0.30} & \multicolumn{1}{l|}{94.48 $\pm$ 0.64} & 92.33 $\pm$ 0.67 & \multicolumn{1}{l|}{94.92 $\pm$ 0.23} & \multicolumn{1}{l|}{95.12 $\pm$ 0.72} & 94.71 $\pm$ 0.67 \\ \hline
\multicolumn{1}{|l|}{\textbf{\textsc{DocEmb} \textsc{FastText} SG}} & \multicolumn{1}{l|}{91.60 $\pm$ 0.49} & \multicolumn{1}{l|}{94.06 $\pm$ 0.54} & 88.81 $\pm$ 0.76 & \multicolumn{1}{l|}{96.00 $\pm$ 0.30} & \multicolumn{1}{l|}{96.48 $\pm$ 0.35} & 95.48 $\pm$ 0.74 \\ \hline
\multicolumn{1}{|l|}{\textbf{\textsc{DocEmb} \textsc{GloVe}}} & \multicolumn{1}{l|}{89.57 $\pm$ 0.50} & \multicolumn{1}{l|}{92.57 $\pm$ 0.60} & 86.04 $\pm$ 1.18 & \multicolumn{1}{l|}{94.05 $\pm$ 0.38} & \multicolumn{1}{l|}{94.29 $\pm$ 0.89} & 93.79 $\pm$ 0.60 \\ \hline
\multicolumn{1}{|l|}{\textbf{\textsc{DocEmb} BERT}} & \multicolumn{1}{l|}{97.09 $\pm$ 0.21} & \multicolumn{1}{l|}{97.50 $\pm$ 0.62} & 96.66 $\pm$ 0.62 & \multicolumn{1}{l|}{98.34 $\pm$ 0.19} & \multicolumn{1}{l|}{98.56 $\pm$ 0.62} & 98.11 $\pm$ 0.56 \\ \hline
\multicolumn{1}{|l|}{\textbf{\textsc{DocEmb} \textsc{RoBERTa}}} & \multicolumn{1}{l|}{96.19 $\pm$ 0.50} & \multicolumn{1}{l|}{96.89 $\pm$ 1.62} & 95.49 $\pm$ 0.99 & \multicolumn{1}{l|}{97.28 $\pm$ 0.55} & \multicolumn{1}{l|}{98.51 $\pm$ 1.32} & 96.04 $\pm$ 0.47 \\ \hline
\multicolumn{1}{|l|}{\textbf{\textsc{DocEmb} BART}} & \multicolumn{1}{l|}{98.57 $\pm$ 0.15} & \multicolumn{1}{l|}{98.71 $\pm$ 0.29} & 98.43 $\pm$ 0.33 & \multicolumn{1}{l|}{98.93 $\pm$ 0.16} & \multicolumn{1}{l|}{99.07 $\pm$ 0.55} & 98.80 $\pm$ 0.39 \\ \hline
 & \multicolumn{3}{c|}{\textbf{LSTM}} & \multicolumn{3}{c|}{\textbf{Bidirectional LSTM}} \\ \hline
\multicolumn{1}{|l|}{\textbf{Vectorization}} & \multicolumn{1}{c|}{\textbf{Accuracy}} & \multicolumn{1}{c|}{\textbf{Precision}} & \multicolumn{1}{c|}{\textbf{Recall}} & \multicolumn{1}{c|}{\textbf{Accuracy}} & \multicolumn{1}{c|}{\textbf{Precision}} & \multicolumn{1}{c|}{\textbf{Recall}} \\ \hline
\multicolumn{1}{|l|}{\textbf{TFIDF}} & \multicolumn{1}{l|}{97.88 $\pm$ 0.23} & \multicolumn{1}{l|}{98.20 $\pm$ 0.19} & 97.55 $\pm$ 0.40 & \multicolumn{1}{l|}{97.84 $\pm$ 0.24} & \multicolumn{1}{l|}{98.21 $\pm$ 0.30} & 97.46 $\pm$ 0.49 \\ \hline
\multicolumn{1}{|l|}{\textbf{\textsc{DocEmb} \textsc{Word2Vec} CBOW}} & \multicolumn{1}{l|}{96.59 $\pm$ 0.36} & \multicolumn{1}{l|}{96.38 $\pm$ 1.05} & 96.84 $\pm$ 1.17 & \multicolumn{1}{l|}{96.89 $\pm$ 0.26} & \multicolumn{1}{l|}{96.89 $\pm$ 0.53} & 96.89 $\pm$ 0.38 \\ \hline
\multicolumn{1}{|l|}{\textbf{\textsc{DocEmb} \textsc{Word2Vec} SG}} & \multicolumn{1}{l|}{96.23 $\pm$ 0.31} & \multicolumn{1}{l|}{96.70 $\pm$ 0.97} & 95.76 $\pm$ 1.32 & \multicolumn{1}{l|}{96.39 $\pm$ 0.37} & \multicolumn{1}{l|}{96.81 $\pm$ 1.30} & 95.98 $\pm$ 1.32 \\ \hline
\multicolumn{1}{|l|}{\textbf{\textsc{DocEmb} \textsc{FastText} CBOW}} & \multicolumn{1}{l|}{96.16 $\pm$ 0.26} & \multicolumn{1}{l|}{96.22 $\pm$ 0.55} & 96.11 $\pm$ 0.85 & \multicolumn{1}{l|}{96.30 $\pm$ 0.32} & \multicolumn{1}{l|}{96.63 $\pm$ 1.19} & 95.98 $\pm$ 1.14 \\ \hline
\multicolumn{1}{|l|}{\textbf{\textsc{DocEmb} \textsc{FastText} SG}} & \multicolumn{1}{l|}{96.61 $\pm$ 0.27} & \multicolumn{1}{l|}{96.52 $\pm$ 0.91} & 96.72 $\pm$ 0.90 & \multicolumn{1}{l|}{96.79 $\pm$ 0.22} & \multicolumn{1}{l|}{96.78 $\pm$ 0.91} & 96.82 $\pm$ 0.88 \\ \hline
\multicolumn{1}{|l|}{\textbf{\textsc{DocEmb} \textsc{GloVe}}} & \multicolumn{1}{l|}{94.66 $\pm$ 0.48} & \multicolumn{1}{l|}{94.92 $\pm$ 2.02} & 94.45 $\pm$ 1.72 & \multicolumn{1}{l|}{94.86 $\pm$ 0.37} & \multicolumn{1}{l|}{95.24 $\pm$ 1.67} & 94.51 $\pm$ 1.63 \\ \hline
\multicolumn{1}{|l|}{\textbf{\textsc{DocEmb} BERT}} & \multicolumn{1}{l|}{98.57 $\pm$ 0.34} & \multicolumn{1}{l|}{98.59 $\pm$ 0.76} & 98.57 $\pm$ 1.10 & \multicolumn{1}{l|}{98.72 $\pm$ 0.40} & \multicolumn{1}{l|}{98.90 $\pm$ 0.81} & 98.55 $\pm$ 0.87 \\ \hline
\multicolumn{1}{|l|}{\textbf{\textsc{DocEmb} \textsc{RoBERTa}}} & \multicolumn{1}{l|}{96.88 $\pm$ 1.57} & \multicolumn{1}{l|}{98.02 $\pm$ 2.95} & 95.80 $\pm$ 1.56 & \multicolumn{1}{l|}{96.97 $\pm$ 1.33} & \multicolumn{1}{l|}{97.78 $\pm$ 2.85} & 96.22 $\pm$ 0.74 \\ \hline
\multicolumn{1}{|l|}{\textbf{\textsc{DocEmb} BART}} & \multicolumn{1}{l|}{99.29 $\pm$ 0.10} & \multicolumn{1}{l|}{99.46 $\pm$ 0.13} & 99.13 $\pm$ 0.14 & \multicolumn{1}{l|}{99.34 $\pm$ 0.08} & \multicolumn{1}{l|}{99.48 $\pm$ 0.12} & 99.20 $\pm$ 0.11 \\ \hline
 & \multicolumn{3}{c|}{\textbf{GRU}} & \multicolumn{3}{c|}{\textbf{Bidirectional GRU}} \\ \hline
\multicolumn{1}{|l|}{\textbf{Vectorization}} & \multicolumn{1}{c|}{\textbf{Accuracy}} & \multicolumn{1}{c|}{\textbf{Precision}} & \multicolumn{1}{c|}{\textbf{Recall}} & \multicolumn{1}{c|}{\textbf{Accuracy}} & \multicolumn{1}{c|}{\textbf{Precision}} & \multicolumn{1}{c|}{\textbf{Recall}} \\ \hline
\multicolumn{1}{|l|}{\textbf{TFIDF}} & \multicolumn{1}{l|}{97.88 $\pm$ 0.24} & \multicolumn{1}{l|}{98.28 $\pm$ 0.27} & 97.47 $\pm$ 0.45 & \multicolumn{1}{l|}{97.84 $\pm$ 0.30} & \multicolumn{1}{l|}{98.04 $\pm$ 0.60} & 97.64 $\pm$ 0.58 \\ \hline
\multicolumn{1}{|l|}{\textbf{\textsc{DocEmb} \textsc{Word2Vec} CBOW}} & \multicolumn{1}{l|}{96.56 $\pm$ 0.29} & \multicolumn{1}{l|}{96.43 $\pm$ 0.75} & 96.71 $\pm$ 1.01 & \multicolumn{1}{l|}{96.57 $\pm$ 0.23} & \multicolumn{1}{l|}{96.37 $\pm$ 0.90} & 96.81 $\pm$ 0.69 \\ \hline
\multicolumn{1}{|l|}{\textbf{\textsc{DocEmb} \textsc{Word2Vec} SG}} & \multicolumn{1}{l|}{96.21 $\pm$ 0.44} & \multicolumn{1}{l|}{95.91 $\pm$ 1.44} & 96.58 $\pm$ 0.97 & \multicolumn{1}{l|}{96.35 $\pm$ 0.38} & \multicolumn{1}{l|}{96.45 $\pm$ 1.09} & 96.26 $\pm$ 1.47 \\ \hline
\multicolumn{1}{|l|}{\textbf{\textsc{DocEmb} \textsc{FastText} CBOW}} & \multicolumn{1}{l|}{96.12 $\pm$ 0.16} & \multicolumn{1}{l|}{96.54 $\pm$ 1.04} & 95.70 $\pm$ 1.00 & \multicolumn{1}{l|}{96.20 $\pm$ 0.33} & \multicolumn{1}{l|}{96.47 $\pm$ 0.88} & 95.91 $\pm$ 1.00 \\ \hline
\multicolumn{1}{|l|}{\textbf{\textsc{DocEmb} \textsc{FastText} SG}} & \multicolumn{1}{l|}{96.40 $\pm$ 0.35} & \multicolumn{1}{l|}{96.11 $\pm$ 1.22} & 96.74 $\pm$ 1.32 & \multicolumn{1}{l|}{96.76 $\pm$ 0.22} & \multicolumn{1}{l|}{96.76 $\pm$ 0.59} & 96.77 $\pm$ 0.50 \\ \hline
\multicolumn{1}{|l|}{\textbf{\textsc{DocEmb} \textsc{GloVe}}} & \multicolumn{1}{l|}{94.62 $\pm$ 0.64} & \multicolumn{1}{l|}{95.60 $\pm$ 2.30} & 93.66 $\pm$ 1.81 & \multicolumn{1}{l|}{94.84 $\pm$ 0.60} & \multicolumn{1}{l|}{95.80 $\pm$ 1.74} & 93.86 $\pm$ 2.01 \\ \hline
\multicolumn{1}{|l|}{\textbf{\textsc{DocEmb} BERT}} & \multicolumn{1}{l|}{98.82 $\pm$ 0.11} & \multicolumn{1}{l|}{98.71 $\pm$ 0.52} & 98.92 $\pm$ 0.48 & \multicolumn{1}{l|}{98.61 $\pm$ 0.41} & \multicolumn{1}{l|}{99.17 $\pm$ 0.47} & 98.05 $\pm$ 1.14 \\ \hline
\multicolumn{1}{|l|}{\textbf{\textsc{DocEmb} \textsc{RoBERTa}}} & \multicolumn{1}{l|}{97.37 $\pm$ 0.60} & \multicolumn{1}{l|}{99.17 $\pm$ 0.43} & 95.56 $\pm$ 1.47 & \multicolumn{1}{l|}{97.31 $\pm$ 0.44} & \multicolumn{1}{l|}{98.34 $\pm$ 1.24} & 96.26 $\pm$ 0.72 \\ \hline
\multicolumn{1}{|l|}{\textbf{\textsc{DocEmb} BART}} & \multicolumn{1}{l|}{99.31 $\pm$ 0.10} & \multicolumn{1}{l|}{99.39 $\pm$ 0.23} & 99.22 $\pm$ 0.14 & \multicolumn{1}{l|}{\textbf{99.36 $\pm$ 0.08}} & \multicolumn{1}{l|}{99.50 $\pm$ 0.09} & 99.22 $\pm$ 0.09 \\ \hline
& \multicolumn{2}{c|}{\textbf{Accuracy}} & \multicolumn{2}{c|}{\textbf{Precision}} & \multicolumn{2}{c|}{\textbf{Recall}} \\ \hline
\multicolumn{1}{|l|}{\textbf{MisRoBÆRTa~\citep{Truica2022}}}   & \multicolumn{2}{c|}{99.34 $\pm$ 0.03}         & \multicolumn{2}{c|}{99.34 $\pm$ 0.03}          & \multicolumn{2}{c|}{99.34 $\pm$ 0.02}       \\ \hline
\end{tabular}
}
\end{table*}

The overall best performance (i.e., an accuracy of $99.36\%$) was obtained with the BiGRU model when employing the document embeddings obtained with BART.
The BiLSTM model obtained similar accuracy as when using BART, i.e., $99.36\%$.
We observe that, regardless of the model, the best results are obtained when employing the document embeddings created with BART.

Regardless of the model, TFIDF obtained very good results.
For some models, it even outperformed the document embeddings obtained from the word or transformer embeddings, e.g., Naïve Bayes with TFIDF obtained an accuracy of $92.69\%$ while Naïve Bayes with \textsc{RoBERTa} obtained an accuracy of $91.98\%$, Gradient Boosted Trees with TFIDF obtained an accuracy of $98.76\%$ while Gradient Boosted Trees with \textsc{RoBERTa} obtained an accuracy of $98.72\%$.
It is worth noting that the worst results were obtained with Naïve Bayes when employing the document embeddings created using the word embeddings.

%MDPI: There is no ``(4)'', please check if the content is missing.
%Authors: We corrected the problem and highlighted the modification
Taking a closer look at the neural networks, we observe that the model ranking when considering accuracy is as follows: (1) BiGRU, (2) BiLSTM, (3) GRU,  (4) LSTM, (5) Multi-Layer Perceptron, and (6) Perceptron.
We note that LSTM and GRU obtained similar results with regard to document embedding, i.e., the results obtained by LSTM with the document embedding employing \textsc{Word2Vec} CBOW are very similar to the results obtained by GRU with the same document embedding model.
The same observation holds when comparing BiLSTM and BiGRU.
Moreover, the neural models that use document embeddings created using word embeddings were outperformed by the ones using transformer embeddings and TFIDF. 
The highest accuracy ($99.36\%$) was obtained by BiGRU with the document embeddings created using BART.

The TFIDF approach proves that the relevance of calculating the importance of each word from a document is an important factor for the fake news detection problem.
This result is a direct consequence of the size of the document-term matrix used as the input.
Moreover, the transformer embeddings obtained the best results among the document embeddings experiments as they manage to encode and preserve the context within the vector representation.
When compared to the state-of-the-art model MisRoBÆRTa~\citep{Truica2022}, the BiLSTM with BART obtained similar results, while the BiGRU with BART marginally outperformed the model with a $0.02\%$ difference in accuracy.
We hypothesize that this difference in performance is due to the use of pre-trained transformers instead of fine-tuned versions. 

The recall metric is the most relevant one for the fake news detection task because it calculates the documents correctly classified as fake relative to all the actual fake documents, regardless of the predicted label.
No clear pattern emerges among the document embeddings to determine which has the overall best performance.
For example, when using LSTM, the best performance was obtained with document embedding \textsc{DocEmb} \textsc{Word2Vec} CBOW ($96.84\%$), followed closely by \textsc{DocEmb} \textsc{FastText} SG ($96.72\%$), while, when using Multi-Layer Perceptron, the best performance was obtained by \textsc{DocEmb} \textsc{FastText} SG ($95.48\%$), followed closely by \textsc{DocEmb} \textsc{Word2Vec} SG ($95.40\%$). 

Finally, by analyzing the results, we observed the following:
\begin{itemize}
    \item[(1)] A simpler neural architecture offers similar or better results compared to complex deep learning architectures that employ multiple layers, i.e., in our comparison, we obtained similar results as the complex MisRoBÆRTa~\citep{Truica2022} architecture without fine-tuning the transformers;
    \item[(2)] The embeddings used to vectorize the textual data make all the difference in performance, i.e., the right embedding must be selected to obtain good results with a given model;
    \item[(3)] We need a data-driven approach to select the best model and the best embedding for our dataset.
\end{itemize}

\subsection{Additional Experiments}

In this section, we present more experiments using four additional datasets that are analyzed in detail in~\citep{DUlizia2021}.
For this set of experiments, we compared our results with existing results from the current literature.
Furthermore, we trained our own model for each dataset using MisRoBÆRTa~\citep{Truica2022}, but we did not fine-tune the transformers.
We used the pre-trained BART (\textit{facebook/bart-large}) and \textsc{RoBERTa} (\textit{roberta-base}) versions from HuggingFace~\citep{Wolf2020}.
We hypothesize that this is the reason we obtained similar results to the ones obtained with the models that use document embeddings with this state-of-the-art architecture.

Tables~\ref{tab:results_liar_multi_full} and~\ref{tab:results_liar_bin_full} present experimental results obtained on the LIAR dataset~\citep{Wang2017}.
For our experiments, we used the dataset as it was initially released, with 6 labels~\citep{Wang2017} (Table~\ref{tab:results_liar_multi_full}), and by balancing the dataset's labels (Table~\ref{tab:results_liar_bin_full}) as proposed in~\citep{Upadhayay2020}.
To balance the labels, we created binary labels, i.e., all the texts that are not labeled with \textit{true} are considered \textit{false}.
Using the same experimental configurations as presented in Section~\ref{subsec:experimental_setup}, we obtained results that are aligned with our original observations on the proposed dataset.
Further, we obtained results similar to state-of-the-art results for the multi-label dataset, e.g., \mbox{\citet{Wang2017}} and \citet{Alhindi2018} obtained an accuracy of $\sim$20\%.
For the binary classification, we obtained results that go beyond the the state of the art, e.g.,\citet{Upadhayay2020} obtains an accuracy of $70\%$ while we obtain an accuracy of $83.99\%$ with the LSTM model that employs the document embeddings constructed with \textsc{GloVe}.

\begin{table*}[!htb]
\centering
\caption{Fake news detection results on Liar dataset with 6 labels as presented in~\citet{Wang2017}}
\label{tab:results_liar_multi_full}
\resizebox{1\textwidth}{!}{%
\begin{tabular}{llll|lll|}
\cline{2-7}
 & \multicolumn{3}{|c|}{\textbf{Naïve Bayes}} & \multicolumn{3}{c|}{\textbf{Gradient   Boosted Trees}} \\ \hline
\multicolumn{1}{|l|}{\textbf{Vectorization}} & \multicolumn{1}{c|}{\textbf{Accuracy}} & \multicolumn{1}{c|}{\textbf{Precision}} & \multicolumn{1}{c|}{\textbf{Recall}} & \multicolumn{1}{c|}{\textbf{Accuracy}} & \multicolumn{1}{c|}{\textbf{Precision}} & \multicolumn{1}{c|}{\textbf{Recall}} \\ \hline
\multicolumn{1}{|l|}{\textbf{TFIDF}} & \multicolumn{1}{l|}{23.07 $\pm$ 0.70} & \multicolumn{1}{l|}{24.16 $\pm$ 2.03} & 23.07 $\pm$ 0.70 & \multicolumn{1}{l|}{23.00 $\pm$    0.93} & \multicolumn{1}{l|}{22.86 $\pm$    0.89} & 23.00 $\pm$    0.93 \\ \hline
\multicolumn{1}{|l|}{\textbf{\textsc{DocEmb} \textsc{Word2Vec} CBOW}} & \multicolumn{1}{l|}{18.32 $\pm$ 1.01} & \multicolumn{1}{l|}{21.62 $\pm$ 1.42} & 18.32 $\pm$ 1.01 & \multicolumn{1}{l|}{22.40 $\pm$  0.59} & \multicolumn{1}{l|}{22.37 $\pm$  0.68} & 22.40 $\pm$  0.59 \\ \hline
\multicolumn{1}{|l|}{\textbf{\textsc{DocEmb} \textsc{Word2Vec} SG}} & \multicolumn{1}{l|}{20.42 $\pm$ 0.96} & \multicolumn{1}{l|}{21.74 $\pm$ 0.84} & 20.42 $\pm$ 0.96 & \multicolumn{1}{l|}{23.12 $\pm$  0.69} & \multicolumn{1}{l|}{23.29 $\pm$  0.69} & 23.12 $\pm$  0.69 \\ \hline
\multicolumn{1}{|l|}{\textbf{\textsc{DocEmb} \textsc{FastText} CBOW}} & \multicolumn{1}{l|}{17.19 $\pm$ 0.63} & \multicolumn{1}{l|}{21.89 $\pm$ 1.45} & 17.19 $\pm$ 0.63 & \multicolumn{1}{l|}{22.68 $\pm$  0.55} & \multicolumn{1}{l|}{22.63 $\pm$  0.71} & 22.68 $\pm$  0.55 \\ \hline
\multicolumn{1}{|l|}{\textbf{\textsc{DocEmb} \textsc{FastText} SG}} & \multicolumn{1}{l|}{19.85 $\pm$ 1.10} & \multicolumn{1}{l|}{21.57 $\pm$ 1.22} & 19.85 $\pm$ 1.10 & \multicolumn{1}{l|}{22.93 $\pm$  0.83} & \multicolumn{1}{l|}{23.00 $\pm$  0.80} & 22.93 $\pm$  0.83 \\ \hline
\multicolumn{1}{|l|}{\textbf{\textsc{DocEmb} \textsc{GloVe}}} & \multicolumn{1}{l|}{17.60 $\pm$ 0.77} & \multicolumn{1}{l|}{21.31 $\pm$ 1.18} & 17.60 $\pm$ 0.77 & \multicolumn{1}{l|}{21.99 $\pm$  0.63} & \multicolumn{1}{l|}{21.72 $\pm$  0.71} & 21.99 $\pm$  0.63 \\ \hline
\multicolumn{1}{|l|}{\textbf{\textsc{DocEmb} BERT}} & \multicolumn{1}{l|}{20.58 $\pm$ 0.71} & \multicolumn{1}{l|}{22.40 $\pm$ 1.17} & 20.58 $\pm$ 0.71 & \multicolumn{1}{l|}{23.78 $\pm$  0.82} & \multicolumn{1}{l|}{24.03 $\pm$  0.97} & 23.78 $\pm$  0.82 \\ \hline
\multicolumn{1}{|l|}{\textbf{\textsc{DocEmb} \textsc{RoBERTa}}} & \multicolumn{1}{l|}{15.91 $\pm$ 1.02} & \multicolumn{1}{l|}{20.31 $\pm$ 1.38} & 15.91 $\pm$ 1.02 & \multicolumn{1}{l|}{21.09 $\pm$  1.08} & \multicolumn{1}{l|}{20.51 $\pm$  0.85} & 21.09 $\pm$  1.08 \\ \hline
\multicolumn{1}{|l|}{\textbf{\textsc{DocEmb} BART}} & \multicolumn{1}{l|}{21.79 $\pm$ 0.90} & \multicolumn{1}{l|}{24.07 $\pm$ 1.09} & 21.79 $\pm$ 0.90 & \multicolumn{1}{l|}{24.93 $\pm$  0.74} & \multicolumn{1}{l|}{25.26 $\pm$  0.83} & 24.93 $\pm$  0.74 \\ \hline
\textbf{} & \multicolumn{3}{|c|}{\textbf{Perceptron}} & \multicolumn{3}{c|}{\textbf{Multi Layer Perceptron}} \\ \hline
\multicolumn{1}{|l|}{\textbf{Vectorization}} & \multicolumn{1}{c|}{\textbf{Accuracy}} & \multicolumn{1}{c|}{\textbf{Precision}} & \multicolumn{1}{c|}{\textbf{Recall}} & \multicolumn{1}{c|}{\textbf{Accuracy}} & \multicolumn{1}{c|}{\textbf{Precision}} & \multicolumn{1}{c|}{\textbf{Recall}} \\ \hline
\multicolumn{1}{|l|}{\textbf{TFIDF}} & \multicolumn{1}{l|}{23.71 $\pm$ 0.68} & \multicolumn{1}{l|}{24.79 $\pm$ 1.26} & 23.71 $\pm$ 0.68 & \multicolumn{1}{l|}{23.04 $\pm$ 0.66} & \multicolumn{1}{l|}{23.08 $\pm$ 0.77} & 23.04 $\pm$ 0.66 \\ \hline
\multicolumn{1}{|l|}{\textbf{\textsc{DocEmb} \textsc{Word2Vec} CBOW}} & \multicolumn{1}{l|}{22.83 $\pm$ 0.65} & \multicolumn{1}{l|}{22.37 $\pm$ 0.66} & 22.83 $\pm$ 0.65 & \multicolumn{1}{l|}{22.70 $\pm$ 0.91} & \multicolumn{1}{l|}{22.56 $\pm$ 0.87} & 22.70 $\pm$ 0.91 \\ \hline
\multicolumn{1}{|l|}{\textbf{\textsc{DocEmb} \textsc{Word2Vec} SG}} & \multicolumn{1}{l|}{23.46 $\pm$ 0.73} & \multicolumn{1}{l|}{23.19 $\pm$ 0.68} & 23.46 $\pm$ 0.73 & \multicolumn{1}{l|}{23.26 $\pm$ 0.65} & \multicolumn{1}{l|}{23.15 $\pm$ 1.09} & 23.26 $\pm$ 0.65 \\ \hline
\multicolumn{1}{|l|}{\textbf{\textsc{DocEmb} \textsc{FastText} CBOW}} & \multicolumn{1}{l|}{22.34 $\pm$ 0.49} & \multicolumn{1}{l|}{21.63 $\pm$ 0.81} & 22.34 $\pm$ 0.49 & \multicolumn{1}{l|}{22.72 $\pm$ 0.89} & \multicolumn{1}{l|}{22.16 $\pm$ 1.19} & 22.72 $\pm$ 0.89 \\ \hline
\multicolumn{1}{|l|}{\textbf{\textsc{DocEmb} \textsc{FastText} SG}} & \multicolumn{1}{l|}{23.62 $\pm$ 0.82} & \multicolumn{1}{l|}{23.29 $\pm$ 1.08} & 23.62 $\pm$ 0.82 & \multicolumn{1}{l|}{23.48 $\pm$ 0.92} & \multicolumn{1}{l|}{23.47 $\pm$ 1.13} & 23.48 $\pm$ 0.92 \\ \hline
\multicolumn{1}{|l|}{\textbf{\textsc{DocEmb} \textsc{GloVe}}} & \multicolumn{1}{l|}{23.64 $\pm$ 0.55} & \multicolumn{1}{l|}{22.54 $\pm$ 0.97} & 23.64 $\pm$ 0.55 & \multicolumn{1}{l|}{23.24 $\pm$ 0.71} & \multicolumn{1}{l|}{21.94 $\pm$ 1.31} & 23.24 $\pm$ 0.71 \\ \hline
\multicolumn{1}{|l|}{\textbf{\textsc{DocEmb} BERT}} & \multicolumn{1}{l|}{24.06 $\pm$ 1.03} & \multicolumn{1}{l|}{24.33 $\pm$ 1.00} & 24.06 $\pm$ 1.03 & \multicolumn{1}{l|}{23.58 $\pm$ 0.66} & \multicolumn{1}{l|}{23.88 $\pm$ 0.84} & 23.58 $\pm$ 0.66 \\ \hline
\multicolumn{1}{|l|}{\textbf{\textsc{DocEmb} \textsc{RoBERTa}}} & \multicolumn{1}{l|}{21.66 $\pm$ 1.59} & \multicolumn{1}{l|}{21.01 $\pm$ 1.60} & 21.66 $\pm$ 1.59 & \multicolumn{1}{l|}{22.82 $\pm$ 0.61} & \multicolumn{1}{l|}{20.83 $\pm$ 1.22} & 22.82 $\pm$ 0.61 \\ \hline
\multicolumn{1}{|l|}{\textbf{\textsc{DocEmb} BART}} & \multicolumn{1}{l|}{25.60 $\pm$ 0.41} & \multicolumn{1}{l|}{25.84 $\pm$ 0.18} & 25.60 $\pm$ 0.41 & \multicolumn{1}{l|}{\textbf{25.89 $\pm$ 0.75}} & \multicolumn{1}{l|}{26.15 $\pm$ 0.72} & 25.89 $\pm$ 0.75 \\ \hline
 & \multicolumn{3}{|c|}{\textbf{LSTM}} & \multicolumn{3}{c|}{\textbf{Bidirectional LSTM}} \\ \hline
\multicolumn{1}{|l|}{\textbf{Vectorization}} & \multicolumn{1}{c|}{\textbf{Accuracy}} & \multicolumn{1}{c|}{\textbf{Precision}} & \multicolumn{1}{c|}{\textbf{Recall}} & \multicolumn{1}{c|}{\textbf{Accuracy}} & \multicolumn{1}{c|}{\textbf{Precision}} & \multicolumn{1}{c|}{\textbf{Recall}} \\ \hline
\multicolumn{1}{|l|}{\textbf{TFIDF}} & \multicolumn{1}{l|}{21.77 $\pm$ 0.50} & \multicolumn{1}{l|}{21.77 $\pm$ 0.48} & 21.77 $\pm$ 0.50 & \multicolumn{1}{l|}{21.62 $\pm$ 0.49} & \multicolumn{1}{l|}{21.60 $\pm$ 0.48} & 21.62 $\pm$ 0.49 \\ \hline
\multicolumn{1}{|l|}{\textbf{\textsc{DocEmb} \textsc{Word2Vec} CBOW}} & \multicolumn{1}{l|}{22.70 $\pm$ 0.95} & \multicolumn{1}{l|}{22.53 $\pm$ 0.94} & 22.70 $\pm$ 0.95 & \multicolumn{1}{l|}{22.65 $\pm$ 0.60} & \multicolumn{1}{l|}{22.57 $\pm$ 0.62} & 22.65 $\pm$ 0.60 \\ \hline
\multicolumn{1}{|l|}{\textbf{\textsc{DocEmb} \textsc{Word2Vec} SG}} & \multicolumn{1}{l|}{23.66 $\pm$ 0.99} & \multicolumn{1}{l|}{23.46 $\pm$ 1.02} & 23.66 $\pm$ 0.99 & \multicolumn{1}{l|}{23.51 $\pm$ 0.69} & \multicolumn{1}{l|}{23.14 $\pm$ 0.84} & 23.51 $\pm$ 0.69 \\ \hline
\multicolumn{1}{|l|}{\textbf{\textsc{DocEmb} \textsc{FastText} CBOW}} & \multicolumn{1}{l|}{22.40 $\pm$ 1.06} & \multicolumn{1}{l|}{22.23 $\pm$ 1.07} & 22.40 $\pm$ 1.06 & \multicolumn{1}{l|}{22.53 $\pm$ 0.85} & \multicolumn{1}{l|}{22.49 $\pm$ 0.86} & 22.53 $\pm$ 0.85 \\ \hline
\multicolumn{1}{|l|}{\textbf{\textsc{DocEmb} \textsc{FastText} SG}} & \multicolumn{1}{l|}{23.50 $\pm$ 0.76} & \multicolumn{1}{l|}{23.24 $\pm$ 0.90} & 23.50 $\pm$ 0.76 & \multicolumn{1}{l|}{23.45 $\pm$ 0.61} & \multicolumn{1}{l|}{23.41 $\pm$ 0.89} & 23.45 $\pm$ 0.61 \\ \hline
\multicolumn{1}{|l|}{\textbf{\textsc{DocEmb} \textsc{GloVe}}} & \multicolumn{1}{l|}{23.59 $\pm$ 0.46} & \multicolumn{1}{l|}{22.90 $\pm$ 1.40} & 23.59 $\pm$ 0.46 & \multicolumn{1}{l|}{23.08 $\pm$ 0.42} & \multicolumn{1}{l|}{22.77 $\pm$ 0.88} & 23.08 $\pm$ 0.42 \\ \hline
\multicolumn{1}{|l|}{\textbf{\textsc{DocEmb} BERT}} & \multicolumn{1}{l|}{23.21 $\pm$ 0.55} & \multicolumn{1}{l|}{23.37 $\pm$ 0.52} & 23.21 $\pm$ 0.55 & \multicolumn{1}{l|}{23.31 $\pm$ 0.70} & \multicolumn{1}{l|}{23.25 $\pm$ 0.76} & 23.31 $\pm$ 0.70 \\ \hline
\multicolumn{1}{|l|}{\textbf{\textsc{DocEmb} \textsc{RoBERTa}}} & \multicolumn{1}{l|}{22.94 $\pm$ 1.00} & \multicolumn{1}{l|}{21.65 $\pm$ 1.08} & 22.94 $\pm$ 1.00 & \multicolumn{1}{l|}{22.96 $\pm$ 0.61} & \multicolumn{1}{l|}{19.77 $\pm$ 1.66} & 22.96 $\pm$ 0.61 \\ \hline
\multicolumn{1}{|l|}{\textbf{\textsc{DocEmb} BART}} & \multicolumn{1}{l|}{25.02 $\pm$ 0.57} & \multicolumn{1}{l|}{25.08 $\pm$ 0.59} & 25.02 $\pm$ 0.57 & \multicolumn{1}{l|}{25.75 $\pm$ 0.61} & \multicolumn{1}{l|}{25.78 $\pm$ 0.62} & 25.75 $\pm$ 0.61 \\ \hline
 & \multicolumn{3}{|c|}{\textbf{GRU}} & \multicolumn{3}{c|}{\textbf{Bidirectional GRU}} \\ \hline
\multicolumn{1}{|l|}{\textbf{Vectorization}} & \multicolumn{1}{c|}{\textbf{Accuracy}} & \multicolumn{1}{c|}{\textbf{Precision}} & \multicolumn{1}{c|}{\textbf{Recall}} & \multicolumn{1}{c|}{\textbf{Accuracy}} & \multicolumn{1}{c|}{\textbf{Precision}} & \multicolumn{1}{c|}{\textbf{Recall}} \\ \hline
\multicolumn{1}{|l|}{\textbf{TFIDF}} & \multicolumn{1}{l|}{21.62 $\pm$ 0.50} & \multicolumn{1}{l|}{21.63 $\pm$ 0.47} & 21.62 $\pm$ 0.50 & \multicolumn{1}{l|}{21.43 $\pm$ 0.44} & \multicolumn{1}{l|}{21.44 $\pm$ 0.46} & 21.43 $\pm$ 0.44 \\ \hline
\multicolumn{1}{|l|}{\textbf{\textsc{DocEmb} \textsc{Word2Vec} CBOW}} & \multicolumn{1}{l|}{22.36 $\pm$ 1.00} & \multicolumn{1}{l|}{22.14 $\pm$ 0.94} & 22.36 $\pm$ 1.00 & \multicolumn{1}{l|}{22.51 $\pm$ 0.76} & \multicolumn{1}{l|}{22.46 $\pm$ 0.79} & 22.51 $\pm$ 0.76 \\ \hline
\multicolumn{1}{|l|}{\textbf{\textsc{DocEmb} \textsc{Word2Vec} SG}} & \multicolumn{1}{l|}{23.47 $\pm$ 0.53} & \multicolumn{1}{l|}{23.02 $\pm$ 0.72} & 23.47 $\pm$ 0.53 & \multicolumn{1}{l|}{23.47 $\pm$ 0.71} & \multicolumn{1}{l|}{23.16 $\pm$ 0.66} & 23.47 $\pm$ 0.71 \\ \hline
\multicolumn{1}{|l|}{\textbf{\textsc{DocEmb} \textsc{FastText} CBOW}} & \multicolumn{1}{l|}{22.62 $\pm$ 0.77} & \multicolumn{1}{l|}{22.48 $\pm$ 0.72} & 22.62 $\pm$ 0.77 & \multicolumn{1}{l|}{22.51 $\pm$ 0.53} & \multicolumn{1}{l|}{22.42 $\pm$ 0.44} & 22.51 $\pm$ 0.53 \\ \hline
\multicolumn{1}{|l|}{\textbf{\textsc{DocEmb} \textsc{FastText} SG}} & \multicolumn{1}{l|}{23.34 $\pm$ 0.55} & \multicolumn{1}{l|}{23.08 $\pm$ 0.76} & 23.34 $\pm$ 0.55 & \multicolumn{1}{l|}{23.54 $\pm$ 0.71} & \multicolumn{1}{l|}{23.28 $\pm$ 0.58} & 23.54 $\pm$ 0.71 \\ \hline
\multicolumn{1}{|l|}{\textbf{\textsc{DocEmb} \textsc{GloVe}}} & \multicolumn{1}{l|}{23.47 $\pm$ 0.64} & \multicolumn{1}{l|}{22.84 $\pm$ 1.76} & 23.47 $\pm$ 0.64 & \multicolumn{1}{l|}{23.21 $\pm$ 0.56} & \multicolumn{1}{l|}{22.93 $\pm$ 1.27} & 23.21 $\pm$ 0.56 \\ \hline
\multicolumn{1}{|l|}{\textbf{\textsc{DocEmb} BERT}} & \multicolumn{1}{l|}{23.64 $\pm$ 0.36} & \multicolumn{1}{l|}{23.90 $\pm$ 0.53} & 23.64 $\pm$ 0.36 & \multicolumn{1}{l|}{23.00 $\pm$ 0.72} & \multicolumn{1}{l|}{23.21 $\pm$ 0.88} & 23.00 $\pm$ 0.72 \\ \hline
\multicolumn{1}{|l|}{\textbf{\textsc{DocEmb} \textsc{RoBERTa}}} & \multicolumn{1}{l|}{22.69 $\pm$ 0.68} & \multicolumn{1}{l|}{19.84 $\pm$ 1.95} & 22.69 $\pm$ 0.68 & \multicolumn{1}{l|}{22.73 $\pm$ 0.72} & \multicolumn{1}{l|}{21.55 $\pm$ 2.26} & 22.73 $\pm$ 0.72 \\ \hline
\multicolumn{1}{|l|}{\textbf{\textsc{DocEmb} BART}} & \multicolumn{1}{l|}{24.99 $\pm$ 0.66} & \multicolumn{1}{l|}{25.00 $\pm$ 0.69} & 24.99 $\pm$ 0.66 & \multicolumn{1}{l|}{25.20 $\pm$ 0.88} & \multicolumn{1}{l|}{25.26 $\pm$ 0.86} & 25.20 $\pm$ 0.88 \\ \hline
\cline{2-7}
& \multicolumn{2}{c|}{\textbf{Accuracy}} & \multicolumn{2}{c|}{\textbf{Precision}} & \multicolumn{2}{c|}{\textbf{Recall}} \\ \hline
\multicolumn{1}{|l|}{\textbf{MisRoBÆRTa~\citep{Truica2022}}}   & \multicolumn{2}{c|}{24.62 $\pm$ 0.39}         & \multicolumn{2}{c|}{25.87 $\pm$ 0.67}          & \multicolumn{2}{c|}{24.61 $\pm$ 0.39}       \\ \hline
& & \multicolumn{5}{|c|}{\textbf{F1-Score}}  \\ \hline
\multicolumn{2}{|l|}{\textbf{Hybrid CNNs~\citep{Wang2017}}}   & \multicolumn{5}{c|}{27.70}        \\ \hline
\multicolumn{2}{|l|}{\textbf{BiLSTM~\citep{Alhindi2018}}}   & \multicolumn{5}{c|}{26.00}        \\ \hline
\end{tabular}
}
\end{table*}

\begin{table*}[!htb]
\centering
\caption{Fake news detection results on Liar dataset with 2 labels as presented in~\citet{Upadhayay2020}}
\label{tab:results_liar_bin_full}
\resizebox{1\textwidth}{!}{%
\begin{tabular}{llll|lll|}
\cline{2-7}
 & \multicolumn{3}{|c|}{\textbf{Naïve Bayes}} & \multicolumn{3}{c|}{\textbf{Gradient   Boosted Trees}} \\ \hline
\multicolumn{1}{|l|}{\textbf{Vectorization}} & \multicolumn{1}{c|}{\textbf{Accuracy}} & \multicolumn{1}{c|}{\textbf{Precision}} & \multicolumn{1}{c|}{\textbf{Recall}} & \multicolumn{1}{c|}{\textbf{Accuracy}} & \multicolumn{1}{c|}{\textbf{Precision}} & \multicolumn{1}{c|}{\textbf{Recall}} \\ \hline
\multicolumn{1}{|l|}{\textbf{TFIDF}} & \multicolumn{1}{l|}{83.92 $\pm$ 0.04} & \multicolumn{1}{l|}{83.96 $\pm$ 0.01} & 99.94 $\pm$ 0.04 & \multicolumn{1}{l|}{83.64 $\pm$ 0.20} & \multicolumn{1}{l|}{84.09 $\pm$ 0.08} & 99.31 $\pm$ 0.19 \\ \hline
\multicolumn{1}{|l|}{\textbf{\textsc{DocEmb} \textsc{Word2Vec} CBOW}} & \multicolumn{1}{l|}{67.02 $\pm$ 1.21} & \multicolumn{1}{l|}{86.09 $\pm$ 0.43} & 72.43 $\pm$ 1.76 & \multicolumn{1}{l|}{83.28 $\pm$ 0.19} & \multicolumn{1}{l|}{84.15 $\pm$ 0.07} & 98.68 $\pm$ 0.25 \\ \hline
\multicolumn{1}{|l|}{\textbf{\textsc{DocEmb} \textsc{Word2Vec} SG}} & \multicolumn{1}{l|}{64.57 $\pm$ 3.54} & \multicolumn{1}{l|}{86.07 $\pm$ 0.31} & 68.96 $\pm$ 5.01 & \multicolumn{1}{l|}{83.30 $\pm$ 0.33} & \multicolumn{1}{l|}{84.15 $\pm$ 0.11} & 98.70 $\pm$ 0.35 \\ \hline
\multicolumn{1}{|l|}{\textbf{\textsc{DocEmb} \textsc{FastText} CBOW}} & \multicolumn{1}{l|}{67.89 $\pm$ 2.19} & \multicolumn{1}{l|}{85.60 $\pm$ 0.32} & 74.26 $\pm$ 3.20 & \multicolumn{1}{l|}{83.25 $\pm$ 0.28} & \multicolumn{1}{l|}{84.16 $\pm$ 0.12} & 98.61 $\pm$ 0.29 \\ \hline
\multicolumn{1}{|l|}{\textbf{\textsc{DocEmb} \textsc{FastText} SG}} & \multicolumn{1}{l|}{65.19 $\pm$ 4.32} & \multicolumn{1}{l|}{86.17 $\pm$ 0.43} & 69.75 $\pm$ 6.17 & \multicolumn{1}{l|}{83.28 $\pm$ 0.23} & \multicolumn{1}{l|}{84.16 $\pm$ 0.10} & 98.65 $\pm$ 0.32 \\ \hline
\multicolumn{1}{|l|}{\textbf{\textsc{DocEmb} \textsc{GloVe}}} & \multicolumn{1}{l|}{59.71 $\pm$ 1.80} & \multicolumn{1}{l|}{85.89 $\pm$ 0.52} & 62.24 $\pm$ 2.54 & \multicolumn{1}{l|}{83.05 $\pm$ 0.26} & \multicolumn{1}{l|}{84.10 $\pm$ 0.08} & 98.42 $\pm$ 0.29 \\ \hline
\multicolumn{1}{|l|}{\textbf{\textsc{DocEmb} BERT}} & \multicolumn{1}{l|}{61.21 $\pm$ 0.91} & \multicolumn{1}{l|}{86.76 $\pm$ 0.58} & 63.49 $\pm$ 1.14 & \multicolumn{1}{l|}{83.38 $\pm$ 0.19} & \multicolumn{1}{l|}{84.16 $\pm$ 0.09} & 98.80 $\pm$ 0.19 \\ \hline
\multicolumn{1}{|l|}{\textbf{\textsc{DocEmb} \textsc{RoBERTa}}} & \multicolumn{1}{l|}{60.04 $\pm$ 3.98} & \multicolumn{1}{l|}{85.30 $\pm$ 0.53} & 63.35 $\pm$ 6.07 & \multicolumn{1}{l|}{83.36 $\pm$ 0.20} & \multicolumn{1}{l|}{84.02 $\pm$ 0.08} & 99.01 $\pm$ 0.25 \\ \hline
\multicolumn{1}{|l|}{\textbf{\textsc{DocEmb} BART}} & \multicolumn{1}{l|}{62.11 $\pm$ 1.03} & \multicolumn{1}{l|}{87.65 $\pm$ 0.52} & 63.87 $\pm$ 1.25 & \multicolumn{1}{l|}{83.46 $\pm$ 0.19} & \multicolumn{1}{l|}{84.36 $\pm$ 0.12} & 98.58 $\pm$ 0.32 \\ \hline
\textbf{} & \multicolumn{3}{|c|}{\textbf{Perceptron}} & \multicolumn{3}{c|}{\textbf{Multi Layer Perceptron}} \\ \hline
\multicolumn{1}{|l|}{\textbf{Vectorization}} & \multicolumn{1}{c|}{\textbf{Accuracy}} & \multicolumn{1}{c|}{\textbf{Precision}} & \multicolumn{1}{c|}{\textbf{Recall}} & \multicolumn{1}{c|}{\textbf{Accuracy}} & \multicolumn{1}{c|}{\textbf{Precision}} & \multicolumn{1}{c|}{\textbf{Recall}} \\ \hline
\multicolumn{1}{|l|}{\textbf{TFIDF}} & \multicolumn{1}{l|}{83.97 $\pm$ 0.01} & \multicolumn{1}{l|}{83.97 $\pm$ 0.01} & 99.99 $\pm$ 0.01 & \multicolumn{1}{l|}{80.87 $\pm$ 0.69} & \multicolumn{1}{l|}{84.50 $\pm$ 0.15} & 94.58 $\pm$ 1.13 \\ \hline
\multicolumn{1}{|l|}{\textbf{\textsc{DocEmb} \textsc{Word2Vec} CBOW}} & \multicolumn{1}{l|}{83.88 $\pm$ 0.06} & \multicolumn{1}{l|}{83.96 $\pm$ 0.02} & 99.88 $\pm$ 0.06 & \multicolumn{1}{l|}{83.96 $\pm$ 0.04} & \multicolumn{1}{l|}{83.98 $\pm$ 0.02} & 99.97 $\pm$ 0.03 \\ \hline
\multicolumn{1}{|l|}{\textbf{\textsc{DocEmb} \textsc{Word2Vec} SG}} & \multicolumn{1}{l|}{83.94 $\pm$ 0.04} & \multicolumn{1}{l|}{83.97 $\pm$ 0.01} & 99.95 $\pm$ 0.06 & \multicolumn{1}{l|}{83.90 $\pm$ 0.10} & \multicolumn{1}{l|}{83.99 $\pm$ 0.04} & 99.86 $\pm$ 0.10 \\ \hline
\multicolumn{1}{|l|}{\textbf{\textsc{DocEmb} \textsc{FastText} CBOW}} & \multicolumn{1}{l|}{83.87 $\pm$ 0.06} & \multicolumn{1}{l|}{83.98 $\pm$ 0.03} & 99.82 $\pm$ 0.08 & \multicolumn{1}{l|}{83.95 $\pm$ 0.06} & \multicolumn{1}{l|}{83.99 $\pm$ 0.03} & 99.94 $\pm$ 0.07 \\ \hline
\multicolumn{1}{|l|}{\textbf{\textsc{DocEmb} \textsc{FastText} SG}} & \multicolumn{1}{l|}{83.97 $\pm$ 0.02} & \multicolumn{1}{l|}{83.97 $\pm$ 0.01} & 99.99 $\pm$ 0.01 & \multicolumn{1}{l|}{83.93 $\pm$ 0.08} & \multicolumn{1}{l|}{83.99 $\pm$ 0.02} & 99.91 $\pm$ 0.10 \\ \hline
\multicolumn{1}{|l|}{\textbf{\textsc{DocEmb} \textsc{GloVe}}} & \multicolumn{1}{l|}{83.97 $\pm$ 0.01} & \multicolumn{1}{l|}{83.97 $\pm$ 0.01} & 99.99 $\pm$ 0.01 & \multicolumn{1}{l|}{83.96 $\pm$ 0.01} & \multicolumn{1}{l|}{83.97 $\pm$ 0.01} & 99.99 $\pm$ 0.01 \\ \hline
\multicolumn{1}{|l|}{\textbf{\textsc{DocEmb} BERT}} & \multicolumn{1}{l|}{83.81 $\pm$ 0.11} & \multicolumn{1}{l|}{83.98 $\pm$ 0.05} & 99.74 $\pm$ 0.14 & \multicolumn{1}{l|}{83.18 $\pm$ 0.52} & \multicolumn{1}{l|}{84.25 $\pm$ 0.25} & 98.37 $\pm$ 1.11 \\ \hline
\multicolumn{1}{|l|}{\textbf{\textsc{DocEmb} \textsc{RoBERTa}}} & \multicolumn{1}{l|}{83.96 $\pm$ 0.03} & \multicolumn{1}{l|}{83.97 $\pm$ 0.01} & 99.99 $\pm$ 0.03 & \multicolumn{1}{l|}{83.97 $\pm$ 0.01} & \multicolumn{1}{l|}{83.97 $\pm$ 0.01} & 99.99 $\pm$ 0.01 \\ \hline
\multicolumn{1}{|l|}{\textbf{\textsc{DocEmb} BART}} & \multicolumn{1}{l|}{83.44 $\pm$ 0.55} & \multicolumn{1}{l|}{84.37 $\pm$ 0.28} & 98.53 $\pm$ 1.17 & \multicolumn{1}{l|}{81.63 $\pm$ 1.01} & \multicolumn{1}{l|}{84.97 $\pm$ 0.33} & 94.91 $\pm$ 1.52 \\ \hline
 & \multicolumn{3}{|c|}{\textbf{LSTM}} & \multicolumn{3}{c|}{\textbf{Bidirectional LSTM}} \\ \hline
\multicolumn{1}{|l|}{\textbf{Vectorization}} & \multicolumn{1}{c|}{\textbf{Accuracy}} & \multicolumn{1}{c|}{\textbf{Precision}} & \multicolumn{1}{c|}{\textbf{Recall}} & \multicolumn{1}{c|}{\textbf{Accuracy}} & \multicolumn{1}{c|}{\textbf{Precision}} & \multicolumn{1}{c|}{\textbf{Recall}} \\ \hline
\multicolumn{1}{|l|}{\textbf{TFIDF}} & \multicolumn{1}{l|}{77.05 $\pm$ 0.89} & \multicolumn{1}{l|}{84.75 $\pm$ 0.20} & 88.62 $\pm$ 1.13 & \multicolumn{1}{l|}{76.96 $\pm$ 0.86} & \multicolumn{1}{l|}{84.75 $\pm$ 0.17} & 88.49 $\pm$ 1.08 \\ \hline
\multicolumn{1}{|l|}{\textbf{\textsc{DocEmb} \textsc{Word2Vec} CBOW}} & \multicolumn{1}{l|}{81.43 $\pm$ 0.57} & \multicolumn{1}{l|}{84.44 $\pm$ 0.13} & 95.47 $\pm$ 0.88 & \multicolumn{1}{l|}{80.51 $\pm$ 1.02} & \multicolumn{1}{l|}{84.63 $\pm$ 0.18} & 93.83 $\pm$ 1.65 \\ \hline
\multicolumn{1}{|l|}{\textbf{\textsc{DocEmb} \textsc{Word2Vec} SG}} & \multicolumn{1}{l|}{83.86 $\pm$ 0.13} & \multicolumn{1}{l|}{84.04 $\pm$ 0.05} & 99.72 $\pm$ 0.13 & \multicolumn{1}{l|}{83.79 $\pm$ 0.16} & \multicolumn{1}{l|}{84.07 $\pm$ 0.05} & 99.57 $\pm$ 0.17 \\ \hline
\multicolumn{1}{|l|}{\textbf{\textsc{DocEmb} \textsc{FastText} CBOW}} & \multicolumn{1}{l|}{81.20 $\pm$ 0.74} & \multicolumn{1}{l|}{84.45 $\pm$ 0.23} & 95.13 $\pm$ 1.03 & \multicolumn{1}{l|}{80.77 $\pm$ 1.12} & \multicolumn{1}{l|}{84.49 $\pm$ 0.34} & 94.44 $\pm$ 1.63 \\ \hline
\multicolumn{1}{|l|}{\textbf{\textsc{DocEmb} \textsc{FastText} SG}} & \multicolumn{1}{l|}{83.88 $\pm$ 0.14} & \multicolumn{1}{l|}{84.01 $\pm$ 0.04} & 99.79 $\pm$ 0.16 & \multicolumn{1}{l|}{83.78 $\pm$ 0.19} & \multicolumn{1}{l|}{84.00 $\pm$ 0.06} & 99.66 $\pm$ 0.19 \\ \hline
\multicolumn{1}{|l|}{\textbf{\textsc{DocEmb} \textsc{GloVe}}} & \multicolumn{1}{l|}{\textbf{83.99 $\pm$ 0.02}} & \multicolumn{1}{l|}{83.98 $\pm$ 0.02} & 99.99 $\pm$ 0.01 & \multicolumn{1}{l|}{83.98 $\pm$ 0.04} & \multicolumn{1}{l|}{83.98 $\pm$ 0.02} & 99.99 $\pm$ 0.01 \\ \hline
\multicolumn{1}{|l|}{\textbf{\textsc{DocEmb} BERT}} & \multicolumn{1}{l|}{80.37 $\pm$ 1.53} & \multicolumn{1}{l|}{84.84 $\pm$ 0.49} & 93.31 $\pm$ 2.74 & \multicolumn{1}{l|}{79.69 $\pm$ 1.76} & \multicolumn{1}{l|}{85.05 $\pm$ 0.35} & 92.00 $\pm$ 2.95 \\ \hline
\multicolumn{1}{|l|}{\textbf{\textsc{DocEmb} \textsc{RoBERTa}}} & \multicolumn{1}{l|}{83.97 $\pm$ 0.01} & \multicolumn{1}{l|}{83.97 $\pm$ 0.01} & 99.99 $\pm$ 0.01 & \multicolumn{1}{l|}{83.97 $\pm$ 0.01} & \multicolumn{1}{l|}{83.97 $\pm$ 0.01} & 99.99 $\pm$ 0.01 \\ \hline
\multicolumn{1}{|l|}{\textbf{\textsc{DocEmb} BART}} & \multicolumn{1}{l|}{81.41 $\pm$ 0.85} & \multicolumn{1}{l|}{84.93 $\pm$ 0.22} & 94.66 $\pm$ 1.21 & \multicolumn{1}{l|}{81.43 $\pm$ 0.94} & \multicolumn{1}{l|}{85.14 $\pm$ 0.22} & 94.34 $\pm$ 1.35 \\ \hline
 & \multicolumn{3}{|c|}{\textbf{GRU}} & \multicolumn{3}{c|}{\textbf{Bidirectional GRU}} \\ \hline
\multicolumn{1}{|l|}{\textbf{Vectorization}} & \multicolumn{1}{c|}{\textbf{Accuracy}} & \multicolumn{1}{c|}{\textbf{Precision}} & \multicolumn{1}{c|}{\textbf{Recall}} & \multicolumn{1}{c|}{\textbf{Accuracy}} & \multicolumn{1}{c|}{\textbf{Precision}} & \multicolumn{1}{c|}{\textbf{Recall}} \\ \hline
\multicolumn{1}{|l|}{\textbf{TFIDF}} & \multicolumn{1}{l|}{76.99 $\pm$ 0.59} & \multicolumn{1}{l|}{84.71 $\pm$ 0.19} & 88.58 $\pm$ 0.65 & \multicolumn{1}{l|}{76.87 $\pm$ 0.66} & \multicolumn{1}{l|}{84.72 $\pm$ 0.22} & 88.39 $\pm$ 0.94 \\ \hline
\multicolumn{1}{|l|}{\textbf{\textsc{DocEmb} \textsc{Word2Vec} CBOW}} & \multicolumn{1}{l|}{81.52 $\pm$ 0.67} & \multicolumn{1}{l|}{84.49 $\pm$ 0.24} & 95.54 $\pm$ 0.95 & \multicolumn{1}{l|}{80.52 $\pm$ 1.01} & \multicolumn{1}{l|}{84.58 $\pm$ 0.15} & 93.92 $\pm$ 1.53 \\ \hline
\multicolumn{1}{|l|}{\textbf{\textsc{DocEmb} \textsc{Word2Vec} SG}} & \multicolumn{1}{l|}{83.91 $\pm$ 0.13} & \multicolumn{1}{l|}{84.04 $\pm$ 0.04} & 99.80 $\pm$ 0.14 & \multicolumn{1}{l|}{83.81 $\pm$ 0.16} & \multicolumn{1}{l|}{84.06 $\pm$ 0.05} & 99.60 $\pm$ 0.18 \\ \hline
\multicolumn{1}{|l|}{\textbf{\textsc{DocEmb} \textsc{FastText} CBOW}} & \multicolumn{1}{l|}{80.93 $\pm$ 1.14} & \multicolumn{1}{l|}{84.44 $\pm$ 0.24} & 94.74 $\pm$ 1.93 & \multicolumn{1}{l|}{80.20 $\pm$ 0.96} & \multicolumn{1}{l|}{84.67 $\pm$ 0.27} & 93.31 $\pm$ 1.55 \\ \hline
\multicolumn{1}{|l|}{\textbf{\textsc{DocEmb} \textsc{FastText} SG}} & \multicolumn{1}{l|}{83.89 $\pm$ 0.12} & \multicolumn{1}{l|}{84.01 $\pm$ 0.04} & 99.80 $\pm$ 0.13 & \multicolumn{1}{l|}{83.82 $\pm$ 0.16} & \multicolumn{1}{l|}{84.02 $\pm$ 0.05} & 99.70 $\pm$ 0.16 \\ \hline
\multicolumn{1}{|l|}{\textbf{\textsc{DocEmb} \textsc{GloVe}}} & \multicolumn{1}{l|}{83.98 $\pm$ 0.03} & \multicolumn{1}{l|}{83.98 $\pm$ 0.02} & 99.99 $\pm$ 0.01 & \multicolumn{1}{l|}{83.98 $\pm$ 0.03} & \multicolumn{1}{l|}{83.99 $\pm$ 0.02} & 99.99 $\pm$ 0.01 \\ \hline
\multicolumn{1}{|l|}{\textbf{\textsc{DocEmb} BERT}} & \multicolumn{1}{l|}{79.86 $\pm$ 2.30} & \multicolumn{1}{l|}{84.73 $\pm$ 0.40} & 92.75 $\pm$ 3.83 & \multicolumn{1}{l|}{79.94 $\pm$ 1.25} & \multicolumn{1}{l|}{84.93 $\pm$ 0.36} & 92.54 $\pm$ 2.08 \\ \hline
\multicolumn{1}{|l|}{\textbf{\textsc{DocEmb} \textsc{RoBERTa}}} & \multicolumn{1}{l|}{83.97 $\pm$ 0.01} & \multicolumn{1}{l|}{83.97 $\pm$ 0.01} & 99.99 $\pm$ 0.01 & \multicolumn{1}{l|}{83.97 $\pm$ 0.01} & \multicolumn{1}{l|}{83.97 $\pm$ 0.01} & 99.99 $\pm$ 0.01 \\ \hline
\multicolumn{1}{|l|}{\textbf{\textsc{DocEmb} BART}} & \multicolumn{1}{l|}{81.22 $\pm$ 0.75} & \multicolumn{1}{l|}{85.13 $\pm$ 0.27} & 94.06 $\pm$ 1.19 & \multicolumn{1}{l|}{80.60 $\pm$ 0.90} & \multicolumn{1}{l|}{85.24 $\pm$ 0.22} & 93.01 $\pm$ 1.34 \\ \hline
\cline{2-7}
& \multicolumn{2}{|c|}{\textbf{Accuracy}} & \multicolumn{2}{c|}{\textbf{Precision}} & \multicolumn{2}{c|}{\textbf{Recall}} \\ \hline
\multicolumn{1}{|l|}{\textbf{MisRoBÆRTa~\citep{Truica2022}}}   & \multicolumn{2}{c|}{ 81.15 $\pm$ 0.07 }         & \multicolumn{2}{c|}{81.15 $\pm$ 0.07}          & \multicolumn{2}{c|}{81.16 $\pm$ 0.07}       \\ \hline
& & & & \multicolumn{3}{|c|}{\textbf{Accuracy}}  \\ \hline
\multicolumn{4}{|l|}{\textbf{CNN with BERT-base embeddings~\citep{Upadhayay2020}}}   & \multicolumn{3}{c|}{70.00}        \\ \hline
\multicolumn{4}{|l|}{\textbf{UFD~\citep{Yang2019}}}   & \multicolumn{3}{c|}{75.90}        \\ \hline
\end{tabular}
}
\end{table*}

Table~\ref{tab:results_liar_multi_full} presents the results obtained by the different machine and deep learning algorithms on the LIAR dataset~\citep{Wang2017}.
The dataset contains approximately 12.8K human annotated short statements collected using \href{https://www.politifact.com/}{POLITIFACT.COM}'s API.
In this set of experiments, we used all the 6 labels of LIAR, i.e., \textit{pants-fire}, \textit{false}, \textit{barely-true}, \textit{half-true}, \textit{mostly-true}, and \textit{true}, to build our classification models.
The dataset is highly imbalanced, as there are more news articles labeled with \textit{true} than news articles labeled with the other five classes combined.
Due to this high degree of imbalance, the models performed poorly.
We observe that the best-performing models employ document embedding constructed with BART.
The overall best performance model was Multi-Layer Perceptron with BART-built document embedding, with an accuracy of $25.89\%$.
The overall difference between the worst- and best-performing models is approximately $7\%$.
We note that for Naïve Bayes, the model trained with TFIDF obtained better scores than the models trained with document embedding.
We observed no real difference in performance among the models trained with the document embeddings using word embeddings.
This low performance is also present in the current literature~\citep{Wang2017,Alhindi2018}, with accuracy scores very similar to the ones obtained by the models we~trained.

To mitigate the poor performance obtained using all 6 labels of the LIAR dataset and to minimize the imbalance between the classes, we employed a binarization approach to the dataset.
This approach is also used in the current literature. 
For example, \citet{Upadhayay2020} and \citet{Yang2019} also use the LIAR dataset with 2 labels, i.e., \textit{true} and \textit{false}, to train their models.
On this dataset, we observed that the performance of all the models improved.
Naïve Bayes trained on document embeddings obtained the worst results.
The overall best results were obtained by LSTM with the document embedding constructed with \textsc{GloVe}, with an accuracy score of $83.99\%$.
Furthermore, Naïve Bayes, Gradient Boosted Trees, and Perceptron obtained better results with the TFIDF vectorization.
The performance of these models is directly impacted by TFIDF's features. 
The proposed approach outperforms more complex models proposed in the current literature, e.g., CNN with BERT-base embeddings~\citep{Upadhayay2020} obtained an accuracy of $70\%$ and UFD~\citep{Yang2019} obtained an accuracy of~$75.90\%$.

Tables~\ref{tab:results_kagle_full}--\ref{tab:results_polifact} present the experimental results obtained on the Kaggle~\citep{Kaliyar2020,Kaliyar2021}, Buzz Feed News~\citep{Horne2017}, and TSHP-17 datasets as presented in~\citep{Rashkin2017,BarronCedeno2019a}.
Both Kaggle and Buzz Feed News are binary datasets, i.e., with the levels reliable and false.
To emphasize that the embedding makes the main difference and that the models can generalize when we move from binary to multi-class classification, we used the multilabel dataset TSHP-17, which has the following 3 classes: satire, hoax, and propaganda.
For this set of experiments, we used the same experimental setup and algorithm configurations as presented above.
Again, we obtained results that are aligned with our original observations, reinforcing our claims.

\begin{table*}[!htb]
\centering
\caption{Fake news detection results on the Kaggle dataset as presented in~\citet{Kaliyar2020,Kaliyar2021}}
\label{tab:results_kagle_full}
\resizebox{1\textwidth}{!}{%
\begin{tabular}{llll|lll|}
\cline{2-7}
 & \multicolumn{3}{|c|}{\textbf{Naïve Bayes}} & \multicolumn{3}{c|}{\textbf{Gradient   Boosted Trees}} \\ \hline
\multicolumn{1}{|l|}{\textbf{Vectorization}} & \multicolumn{1}{c|}{\textbf{Accuracy}} & \multicolumn{1}{c|}{\textbf{Precision}} & \multicolumn{1}{c|}{\textbf{Recall}} & \multicolumn{1}{c|}{\textbf{Accuracy}} & \multicolumn{1}{c|}{\textbf{Precision}} & \multicolumn{1}{c|}{\textbf{Recall}} \\ \hline
\multicolumn{1}{|l|}{\textbf{TFIDF}} & \multicolumn{1}{l|}{89.54 $\pm$ 0.20} & \multicolumn{1}{l|}{92.94 $\pm$ 0.33} & 84.93 $\pm$ 0.67 & \multicolumn{1}{l|}{96.74 $\pm$ 0.15} & \multicolumn{1}{l|}{96.22 $\pm$ 0.36} & 97.10 $\pm$ 0.29 \\ \hline
\multicolumn{1}{|l|}{\textbf{\textsc{DocEmb} \textsc{Word2Vec} CBOW}} & \multicolumn{1}{l|}{71.14 $\pm$ 0.41} & \multicolumn{1}{l|}{78.89 $\pm$ 0.63} & 55.41 $\pm$ 0.62 & \multicolumn{1}{l|}{91.90 $\pm$ 0.23} & \multicolumn{1}{l|}{92.01 $\pm$ 0.46} & 91.24 $\pm$ 0.40 \\ \hline
\multicolumn{1}{|l|}{\textbf{\textsc{DocEmb} \textsc{Word2Vec} SG}} & \multicolumn{1}{l|}{60.91 $\pm$ 0.53} & \multicolumn{1}{l|}{85.19 $\pm$ 1.69} & 23.66 $\pm$ 1.16 & \multicolumn{1}{l|}{93.31 $\pm$ 0.37} & \multicolumn{1}{l|}{93.81 $\pm$ 0.32} & 92.33 $\pm$ 0.55 \\ \hline
\multicolumn{1}{|l|}{\textbf{\textsc{DocEmb} \textsc{FastText} CBOW}} & \multicolumn{1}{l|}{67.45 $\pm$ 0.67} & \multicolumn{1}{l|}{77.26 $\pm$ 1.05} & 46.75 $\pm$ 1.06 & \multicolumn{1}{l|}{91.45 $\pm$ 0.29} & \multicolumn{1}{l|}{91.32 $\pm$ 0.68} & 91.06 $\pm$ 0.49 \\ \hline
\multicolumn{1}{|l|}{\textbf{\textsc{DocEmb} \textsc{FastText} SG}} & \multicolumn{1}{l|}{60.22 $\pm$ 0.22} & \multicolumn{1}{l|}{88.13 $\pm$ 1.08} & 20.93 $\pm$ 0.41 & \multicolumn{1}{l|}{93.41 $\pm$ 0.31} & \multicolumn{1}{l|}{94.05 $\pm$ 0.34} & 92.28 $\pm$ 0.58 \\ \hline
\multicolumn{1}{|l|}{\textbf{\textsc{DocEmb} \textsc{GloVe}}} & \multicolumn{1}{l|}{62.22 $\pm$ 0.49} & \multicolumn{1}{l|}{81.02 $\pm$ 1.27} & 29.04 $\pm$ 0.91 & \multicolumn{1}{l|}{90.63 $\pm$ 0.30} & \multicolumn{1}{l|}{89.98 $\pm$ 0.48} & 90.83 $\pm$ 0.36 \\ \hline
\multicolumn{1}{|l|}{\textbf{\textsc{DocEmb} BERT}} & \multicolumn{1}{l|}{70.52 $\pm$ 0.47} & \multicolumn{1}{l|}{82.03 $\pm$ 0.70} & 50.34 $\pm$ 1.06 & \multicolumn{1}{l|}{92.91 $\pm$ 0.34} & \multicolumn{1}{l|}{93.58 $\pm$ 0.22} & 91.69 $\pm$ 0.65 \\ \hline
\multicolumn{1}{|l|}{\textbf{\textsc{DocEmb} \textsc{RoBERTa}}} & \multicolumn{1}{l|}{81.86 $\pm$ 0.56} & \multicolumn{1}{l|}{87.45 $\pm$ 0.78} & 73.18 $\pm$ 1.27 & \multicolumn{1}{l|}{92.42 $\pm$ 0.22} & \multicolumn{1}{l|}{92.51 $\pm$ 0.33} & 91.82 $\pm$ 0.45 \\ \hline
\multicolumn{1}{|l|}{\textbf{\textsc{DocEmb} BART}} & \multicolumn{1}{l|}{90.14 $\pm$ 0.42} & \multicolumn{1}{l|}{91.69 $\pm$ 0.63} & 87.64 $\pm$ 0.52 & \multicolumn{1}{l|}{99.06 $\pm$ 0.11} & \multicolumn{1}{l|}{98.82 $\pm$ 0.24} & 99.24 $\pm$ 0.19 \\ \hline
\textbf{} & \multicolumn{3}{|c|}{\textbf{Perceptron}} & \multicolumn{3}{c|}{\textbf{Multi Layer Perceptron}} \\ \hline
\multicolumn{1}{|l|}{\textbf{Vectorization}} & \multicolumn{1}{c|}{\textbf{Accuracy}} & \multicolumn{1}{c|}{\textbf{Precision}} & \multicolumn{1}{c|}{\textbf{Recall}} & \multicolumn{1}{c|}{\textbf{Accuracy}} & \multicolumn{1}{c|}{\textbf{Precision}} & \multicolumn{1}{c|}{\textbf{Recall}} \\ \hline
\multicolumn{1}{|l|}{\textbf{TFIDF}} & \multicolumn{1}{l|}{93.76 $\pm$ 0.27} & \multicolumn{1}{l|}{94.45 $\pm$ 0.51} & 92.59 $\pm$ 0.56 & \multicolumn{1}{l|}{95.36 $\pm$ 0.23} & \multicolumn{1}{l|}{95.24 $\pm$ 0.42} & 95.20 $\pm$ 0.44 \\ \hline
\multicolumn{1}{|l|}{\textbf{\textsc{DocEmb} \textsc{Word2Vec} CBOW}} & \multicolumn{1}{l|}{90.15 $\pm$ 0.35} & \multicolumn{1}{l|}{90.84 $\pm$ 0.63} & 88.67 $\pm$ 0.71 & \multicolumn{1}{l|}{91.54 $\pm$ 0.40} & \multicolumn{1}{l|}{91.67 $\pm$ 1.42} & 90.89 $\pm$ 1.45 \\ \hline
\multicolumn{1}{|l|}{\textbf{\textsc{DocEmb} \textsc{Word2Vec} SG}} & \multicolumn{1}{l|}{88.98 $\pm$ 0.46} & \multicolumn{1}{l|}{90.48 $\pm$ 0.51} & 86.41 $\pm$ 0.86 & \multicolumn{1}{l|}{92.45 $\pm$ 0.32} & \multicolumn{1}{l|}{92.59 $\pm$ 1.19} & 91.83 $\pm$ 1.29 \\ \hline
\multicolumn{1}{|l|}{\textbf{\textsc{DocEmb} \textsc{FastText} CBOW}} & \multicolumn{1}{l|}{90.10 $\pm$ 0.44} & \multicolumn{1}{l|}{90.23 $\pm$ 0.95} & 89.30 $\pm$ 0.62 & \multicolumn{1}{l|}{92.00 $\pm$ 0.44} & \multicolumn{1}{l|}{91.99 $\pm$ 1.03} & 91.54 $\pm$ 1.68 \\ \hline
\multicolumn{1}{|l|}{\textbf{\textsc{DocEmb} \textsc{FastText} SG}} & \multicolumn{1}{l|}{88.45 $\pm$ 0.59} & \multicolumn{1}{l|}{90.65 $\pm$ 0.82} & 84.99 $\pm$ 1.07 & \multicolumn{1}{l|}{92.15 $\pm$ 0.43} & \multicolumn{1}{l|}{92.78 $\pm$ 1.03} & 90.94 $\pm$ 0.81 \\ \hline
\multicolumn{1}{|l|}{\textbf{\textsc{DocEmb} \textsc{GloVe}}} & \multicolumn{1}{l|}{83.90 $\pm$ 0.52} & \multicolumn{1}{l|}{85.26 $\pm$ 1.15} & 80.89 $\pm$ 1.97 & \multicolumn{1}{l|}{87.57 $\pm$ 0.57} & \multicolumn{1}{l|}{88.79 $\pm$ 2.16} & 85.29 $\pm$ 2.64 \\ \hline
\multicolumn{1}{|l|}{\textbf{\textsc{DocEmb} BERT}} & \multicolumn{1}{l|}{92.09 $\pm$ 0.50} & \multicolumn{1}{l|}{92.84 $\pm$ 1.01} & 90.74 $\pm$ 1.36 & \multicolumn{1}{l|}{94.80 $\pm$ 0.47} & \multicolumn{1}{l|}{94.51 $\pm$ 1.95} & 94.89 $\pm$ 1.73 \\ \hline
\multicolumn{1}{|l|}{\textbf{\textsc{DocEmb} \textsc{RoBERTa}}} & \multicolumn{1}{l|}{91.62 $\pm$ 0.40} & \multicolumn{1}{l|}{91.02 $\pm$ 2.03} & 91.92 $\pm$ 1.94 & \multicolumn{1}{l|}{92.74 $\pm$ 0.96} & \multicolumn{1}{l|}{93.03 $\pm$ 3.76} & 92.31 $\pm$ 4.31 \\ \hline
\multicolumn{1}{|l|}{\textbf{\textsc{DocEmb} BART}} & \multicolumn{1}{l|}{99.73 $\pm$ 0.09} & \multicolumn{1}{l|}{99.66 $\pm$ 0.11} & 99.78 $\pm$ 0.13 & \multicolumn{1}{l|}{99.77 $\pm$ 0.07} & \multicolumn{1}{l|}{99.70 $\pm$ 0.14} & 99.82 $\pm$ 0.08 \\ \hline
 & \multicolumn{3}{|c|}{\textbf{LSTM}} & \multicolumn{3}{c|}{\textbf{Bidirectional LSTM}} \\ \hline
\multicolumn{1}{|l|}{\textbf{Vectorization}} & \multicolumn{1}{c|}{\textbf{Accuracy}} & \multicolumn{1}{c|}{\textbf{Precision}} & \multicolumn{1}{c|}{\textbf{Recall}} & \multicolumn{1}{c|}{\textbf{Accuracy}} & \multicolumn{1}{c|}{\textbf{Precision}} & \multicolumn{1}{c|}{\textbf{Recall}} \\ \hline
\multicolumn{1}{|l|}{\textbf{TFIDF}} & \multicolumn{1}{l|}{95.05 $\pm$ 0.25} & \multicolumn{1}{l|}{94.83 $\pm$ 0.44} & 95.00 $\pm$ 0.50 & \multicolumn{1}{l|}{95.01 $\pm$ 0.25} & \multicolumn{1}{l|}{94.75 $\pm$ 0.39} & 95.00 $\pm$ 0.34 \\ \hline
\multicolumn{1}{|l|}{\textbf{\textsc{DocEmb} \textsc{Word2Vec} CBOW}} & \multicolumn{1}{l|}{93.70 $\pm$ 0.32} & \multicolumn{1}{l|}{94.11 $\pm$ 1.11} & 92.87 $\pm$ 1.50 & \multicolumn{1}{l|}{93.81 $\pm$ 0.39} & \multicolumn{1}{l|}{93.60 $\pm$ 0.82} & 93.68 $\pm$ 0.87 \\ \hline
\multicolumn{1}{|l|}{\textbf{\textsc{DocEmb} \textsc{Word2Vec} SG}} & \multicolumn{1}{l|}{93.03 $\pm$ 0.32} & \multicolumn{1}{l|}{93.22 $\pm$ 1.35} & 92.40 $\pm$ 1.24 & \multicolumn{1}{l|}{93.14 $\pm$ 0.50} & \multicolumn{1}{l|}{94.08 $\pm$ 1.74} & 91.71 $\pm$ 1.80 \\ \hline
\multicolumn{1}{|l|}{\textbf{\textsc{DocEmb} \textsc{FastText} CBOW}} & \multicolumn{1}{l|}{93.25 $\pm$ 0.65} & \multicolumn{1}{l|}{92.83 $\pm$ 2.21} & 93.44 $\pm$ 2.16 & \multicolumn{1}{l|}{93.52 $\pm$ 0.30} & \multicolumn{1}{l|}{93.04 $\pm$ 1.66} & 93.73 $\pm$ 2.08 \\ \hline
\multicolumn{1}{|l|}{\textbf{\textsc{DocEmb} \textsc{FastText} SG}} & \multicolumn{1}{l|}{92.90 $\pm$ 0.41} & \multicolumn{1}{l|}{93.13 $\pm$ 1.17} & 92.21 $\pm$ 1.15 & \multicolumn{1}{l|}{92.67 $\pm$ 0.56} & \multicolumn{1}{l|}{94.61 $\pm$ 1.65} & 90.12 $\pm$ 2.34 \\ \hline
\multicolumn{1}{|l|}{\textbf{\textsc{DocEmb} \textsc{GloVe}}} & \multicolumn{1}{l|}{88.84 $\pm$ 0.82} & \multicolumn{1}{l|}{88.36 $\pm$ 2.96} & 88.98 $\pm$ 2.98 & \multicolumn{1}{l|}{88.83 $\pm$ 0.79} & \multicolumn{1}{l|}{89.23 $\pm$ 3.27} & 87.93 $\pm$ 4.70 \\ \hline
\multicolumn{1}{|l|}{\textbf{\textsc{DocEmb} BERT}} & \multicolumn{1}{l|}{96.31 $\pm$ 0.72} & \multicolumn{1}{l|}{96.15 $\pm$ 2.41} & 96.36 $\pm$ 1.84 & \multicolumn{1}{l|}{96.36 $\pm$ 1.32} & \multicolumn{1}{l|}{96.08 $\pm$ 3.10} & 96.59 $\pm$ 1.13 \\ \hline
\multicolumn{1}{|l|}{\textbf{\textsc{DocEmb} \textsc{RoBERTa}}} & \multicolumn{1}{l|}{93.52 $\pm$ 0.78} & \multicolumn{1}{l|}{94.41 $\pm$ 2.02} & 92.24 $\pm$ 2.94 & \multicolumn{1}{l|}{93.12 $\pm$ 1.29} & \multicolumn{1}{l|}{94.87 $\pm$ 2.53} & 90.93 $\pm$ 4.62 \\ \hline
\multicolumn{1}{|l|}{\textbf{\textsc{DocEmb} BART}} & \multicolumn{1}{l|}{99.79 $\pm$ 0.06} & \multicolumn{1}{l|}{99.74 $\pm$ 0.11} & 99.84 $\pm$ 0.08 & \multicolumn{1}{l|}{\textbf{99.80 $\pm$ 0.12}} & \multicolumn{1}{l|}{99.72 $\pm$ 0.18} & 99.87 $\pm$ 0.11 \\ \hline
 & \multicolumn{3}{|c|}{\textbf{GRU}} & \multicolumn{3}{c|}{\textbf{Bidirectional GRU}} \\ \hline
\multicolumn{1}{|l|}{\textbf{Vectorization}} & \multicolumn{1}{c|}{\textbf{Accuracy}} & \multicolumn{1}{c|}{\textbf{Precision}} & \multicolumn{1}{c|}{\textbf{Recall}} & \multicolumn{1}{c|}{\textbf{Accuracy}} & \multicolumn{1}{c|}{\textbf{Precision}} & \multicolumn{1}{c|}{\textbf{Recall}} \\ \hline
\multicolumn{1}{|l|}{\textbf{TFIDF}} & \multicolumn{1}{l|}{94.99 $\pm$ 0.22} & \multicolumn{1}{l|}{94.64 $\pm$ 0.34} & 95.08 $\pm$ 0.45 & \multicolumn{1}{l|}{94.98 $\pm$ 0.27} & \multicolumn{1}{l|}{94.71 $\pm$ 0.38} & 94.98 $\pm$ 0.38 \\ \hline
\multicolumn{1}{|l|}{\textbf{\textsc{DocEmb} \textsc{Word2Vec} CBOW}} & \multicolumn{1}{l|}{93.67 $\pm$ 0.41} & \multicolumn{1}{l|}{93.66 $\pm$ 1.02} & 93.32 $\pm$ 1.19 & \multicolumn{1}{l|}{93.68 $\pm$ 0.30} & \multicolumn{1}{l|}{93.48 $\pm$ 0.67} & 93.54 $\pm$ 1.20 \\ \hline
\multicolumn{1}{|l|}{\textbf{\textsc{DocEmb} \textsc{Word2Vec} SG}} & \multicolumn{1}{l|}{93.05 $\pm$ 0.36} & \multicolumn{1}{l|}{93.28 $\pm$ 1.48} & 92.40 $\pm$ 1.35 & \multicolumn{1}{l|}{92.91 $\pm$ 0.77} & \multicolumn{1}{l|}{93.55 $\pm$ 2.49} & 91.87 $\pm$ 2.32 \\ \hline
\multicolumn{1}{|l|}{\textbf{\textsc{DocEmb} \textsc{FastText} CBOW}} & \multicolumn{1}{l|}{93.18 $\pm$ 0.54} & \multicolumn{1}{l|}{92.94 $\pm$ 2.21} & 93.14 $\pm$ 1.85 & \multicolumn{1}{l|}{93.34 $\pm$ 0.59} & \multicolumn{1}{l|}{92.43 $\pm$ 1.26} & 94.02 $\pm$ 1.01 \\ \hline
\multicolumn{1}{|l|}{\textbf{\textsc{DocEmb} \textsc{FastText} SG}} & \multicolumn{1}{l|}{92.73 $\pm$ 0.76} & \multicolumn{1}{l|}{92.94 $\pm$ 2.54} & 92.16 $\pm$ 2.16 & \multicolumn{1}{l|}{92.86 $\pm$ 0.45} & \multicolumn{1}{l|}{94.07 $\pm$ 1.14} & 91.09 $\pm$ 2.10 \\ \hline
\multicolumn{1}{|l|}{\textbf{\textsc{DocEmb} \textsc{GloVe}}} & \multicolumn{1}{l|}{88.83 $\pm$ 0.76} & \multicolumn{1}{l|}{89.73 $\pm$ 3.25} & 87.26 $\pm$ 3.54 & \multicolumn{1}{l|}{89.19 $\pm$ 0.64} & \multicolumn{1}{l|}{90.81 $\pm$ 1.47} & 86.59 $\pm$ 2.85 \\ \hline
\multicolumn{1}{|l|}{\textbf{\textsc{DocEmb} BERT}} & \multicolumn{1}{l|}{96.25 $\pm$ 0.52} & \multicolumn{1}{l|}{96.81 $\pm$ 1.95} & 95.50 $\pm$ 1.87 & \multicolumn{1}{l|}{96.37 $\pm$ 0.81} & \multicolumn{1}{l|}{97.71 $\pm$ 0.90} & 94.77 $\pm$ 2.28 \\ \hline
\multicolumn{1}{|l|}{\textbf{\textsc{DocEmb} \textsc{RoBERTa}}} & \multicolumn{1}{l|}{92.32 $\pm$ 1.93} & \multicolumn{1}{l|}{92.10 $\pm$ 5.31} & 92.80 $\pm$ 5.35 & \multicolumn{1}{l|}{92.93 $\pm$ 1.27} & \multicolumn{1}{l|}{95.36 $\pm$ 2.48} & 90.01 $\pm$ 4.51 \\ \hline
\multicolumn{1}{|l|}{\textbf{\textsc{DocEmb} BART}} & \multicolumn{1}{l|}{99.77 $\pm$ 0.07} & \multicolumn{1}{l|}{99.72 $\pm$ 0.14} & 99.82 $\pm$ 0.09 & \multicolumn{1}{l|}{99.78 $\pm$ 0.07} & \multicolumn{1}{l|}{99.72 $\pm$ 0.12} & 99.83 $\pm$ 0.09 \\ \hline
& & & & \multicolumn{1}{|c|}{\textbf{Accuracy}} & \multicolumn{1}{c|}{\textbf{Precision}} & \multicolumn{1}{c|}{\textbf{Recall}} \\ \hline
\multicolumn{4}{|l|}{\textbf{MisRoBÆRTa~\citep{Truica2022}}}   & \multicolumn{1}{c|}{ 97.57 $\pm$ 0.29 }         & \multicolumn{1}{c|}{97.58 $\pm$ 0.28}          & \multicolumn{1}{c|}{97.57 $\pm$ 0.31}       \\ \hline
\multicolumn{4}{|l|}{\textbf{C-CNN~\citep{Sedik2022}}}   & \multicolumn{1}{c|}{ 99.90 }         & \multicolumn{1}{c|}{99.90}          & \multicolumn{1}{c|}{99.90 }       \\ \hline

& & \multicolumn{5}{|c|}{\textbf{Accuracy}}  \\ \hline
\multicolumn{2}{|l|}{\textbf{FNDNet~\citep{Kaliyar2020}}}   & \multicolumn{5}{c|}{98.36}        \\ \hline
\multicolumn{2}{|l|}{\textbf{FakeBERT~\citep{Kaliyar2021}}}   & \multicolumn{5}{c|}{98.90}        \\ \hline

\end{tabular}
}
\end{table*}

\begin{table*}[!htb]
\centering
\caption{Fake news detection results on the Buzz Feed News dataset as presented in~\citet{Horne2017}}
\label{tab:results_buzzfeed}
\resizebox{1\textwidth}{!}{%
\begin{tabular}{llll|lll|}
\cline{2-7}
 & \multicolumn{3}{|c|}{\textbf{Naïve Bayes}} & \multicolumn{3}{c|}{\textbf{Gradient   Boosted Trees}} \\ \hline
\multicolumn{1}{|l|}{\textbf{Vectorization}} & \multicolumn{1}{c|}{\textbf{Accuracy}} & \multicolumn{1}{c|}{\textbf{Precision}} & \multicolumn{1}{c|}{\textbf{Recall}} & \multicolumn{1}{c|}{\textbf{Accuracy}} & \multicolumn{1}{c|}{\textbf{Precision}} & \multicolumn{1}{c|}{\textbf{Recall}} \\ \hline
\multicolumn{1}{|l|}{\textbf{TFIDF}} & \multicolumn{1}{l|}{77.88 $\pm$ 0.01} & \multicolumn{1}{l|}{60.66 $\pm$ 0.01} & 77.88 $\pm$ 0.01 & \multicolumn{1}{l|}{78.29 $\pm$ 0.99} & \multicolumn{1}{l|}{70.70 $\pm$ 2.62} & 78.29 $\pm$ 0.99 \\ \hline
\multicolumn{1}{|l|}{\textbf{\textsc{DocEmb} \textsc{Word2Vec} CBOW}} & \multicolumn{1}{l|}{55.26 $\pm$ 1.52} & \multicolumn{1}{l|}{72.29 $\pm$ 2.16} & 55.26 $\pm$ 1.52 & \multicolumn{1}{l|}{78.72 $\pm$ 0.88} & \multicolumn{1}{l|}{72.97 $\pm$ 2.71} & 78.72 $\pm$ 0.88 \\ \hline
\multicolumn{1}{|l|}{\textbf{\textsc{DocEmb} \textsc{Word2Vec} SG}} & \multicolumn{1}{l|}{54.42 $\pm$ 2.99} & \multicolumn{1}{l|}{72.24 $\pm$ 2.05} & 54.42 $\pm$ 2.99 & \multicolumn{1}{l|}{78.97 $\pm$ 0.91} & \multicolumn{1}{l|}{73.30 $\pm$ 2.66} & 78.97 $\pm$ 0.91 \\ \hline
\multicolumn{1}{|l|}{\textbf{\textsc{DocEmb} \textsc{FastText} CBOW}} & \multicolumn{1}{l|}{49.13 $\pm$ 2.26} & \multicolumn{1}{l|}{71.76 $\pm$ 1.86} & 49.13 $\pm$ 2.26 & \multicolumn{1}{l|}{78.54 $\pm$ 0.57} & \multicolumn{1}{l|}{72.26 $\pm$ 1.50} & 78.54 $\pm$ 0.57 \\ \hline
\multicolumn{1}{|l|}{\textbf{\textsc{DocEmb} \textsc{FastText} SG}} & \multicolumn{1}{l|}{55.58 $\pm$ 2.35} & \multicolumn{1}{l|}{73.95 $\pm$ 1.76} & 55.58 $\pm$ 2.35 & \multicolumn{1}{l|}{78.94 $\pm$ 0.66} & \multicolumn{1}{l|}{73.46 $\pm$ 2.97} & 78.94 $\pm$ 0.66 \\ \hline
\multicolumn{1}{|l|}{\textbf{\textsc{DocEmb} \textsc{GloVe}}} & \multicolumn{1}{l|}{48.63 $\pm$ 8.40} & \multicolumn{1}{l|}{69.79 $\pm$ 3.13} & 48.63 $\pm$ 8.40 & \multicolumn{1}{l|}{78.16 $\pm$ 1.35} & \multicolumn{1}{l|}{71.89 $\pm$ 4.01} & 78.16 $\pm$ 1.35 \\ \hline
\multicolumn{1}{|l|}{\textbf{\textsc{DocEmb} BERT}} & \multicolumn{1}{l|}{59.91 $\pm$ 1.79} & \multicolumn{1}{l|}{76.22 $\pm$ 0.85} & 59.91 $\pm$ 1.79 & \multicolumn{1}{l|}{78.44 $\pm$ 0.73} & \multicolumn{1}{l|}{71.31 $\pm$ 1.95} & 78.44 $\pm$ 0.73 \\ \hline
\multicolumn{1}{|l|}{\textbf{\textsc{DocEmb} \textsc{RoBERTa}}} & \multicolumn{1}{l|}{62.02 $\pm$ 7.65} & \multicolumn{1}{l|}{70.54 $\pm$ 1.52} & 62.02 $\pm$ 7.65 & \multicolumn{1}{l|}{77.98 $\pm$ 0.70} & \multicolumn{1}{l|}{69.83 $\pm$ 4.17} & 77.98 $\pm$ 0.70 \\ \hline
\multicolumn{1}{|l|}{\textbf{\textsc{DocEmb} BART}} & \multicolumn{1}{l|}{61.56 $\pm$ 1.29} & \multicolumn{1}{l|}{80.57 $\pm$ 1.17} & 61.56 $\pm$ 1.29 & \multicolumn{1}{l|}{79.28 $\pm$ 1.18} & \multicolumn{1}{l|}{73.92 $\pm$ 2.50} & 79.28 $\pm$ 1.18 \\ \hline
\textbf{} & \multicolumn{3}{|c|}{\textbf{Perceptron}} & \multicolumn{3}{c|}{\textbf{Multi Layer Perceptron}} \\ \hline
\multicolumn{1}{|l|}{\textbf{Vectorization}} & \multicolumn{1}{c|}{\textbf{Accuracy}} & \multicolumn{1}{c|}{\textbf{Precision}} & \multicolumn{1}{c|}{\textbf{Recall}} & \multicolumn{1}{c|}{\textbf{Accuracy}} & \multicolumn{1}{c|}{\textbf{Precision}} & \multicolumn{1}{c|}{\textbf{Recall}} \\ \hline
\multicolumn{1}{|l|}{\textbf{TFIDF}} & \multicolumn{1}{l|}{77.88 $\pm$ 0.01} & \multicolumn{1}{l|}{60.66 $\pm$ 0.01} & 77.88 $\pm$ 0.01 & \multicolumn{1}{l|}{78.29 $\pm$ 0.50} & \multicolumn{1}{l|}{70.91 $\pm$ 6.32} & 78.29 $\pm$ 0.50 \\ \hline
\multicolumn{1}{|l|}{\textbf{\textsc{DocEmb} \textsc{Word2Vec} CBOW}} & \multicolumn{1}{l|}{77.98 $\pm$ 0.31} & \multicolumn{1}{l|}{63.63 $\pm$ 3.26} & 77.98 $\pm$ 0.31 & \multicolumn{1}{l|}{78.10 $\pm$ 0.50} & \multicolumn{1}{l|}{65.94 $\pm$ 5.52} & 78.10 $\pm$ 0.50 \\ \hline
\multicolumn{1}{|l|}{\textbf{\textsc{DocEmb} \textsc{Word2Vec} SG}} & \multicolumn{1}{l|}{77.88 $\pm$ 0.01} & \multicolumn{1}{l|}{60.66 $\pm$ 0.01} & 77.88 $\pm$ 0.01 & \multicolumn{1}{l|}{77.88 $\pm$ 0.01} & \multicolumn{1}{l|}{60.66 $\pm$ 0.01} & 77.88 $\pm$ 0.01 \\ \hline
\multicolumn{1}{|l|}{\textbf{\textsc{DocEmb} \textsc{FastText} CBOW}} & \multicolumn{1}{l|}{77.79 $\pm$ 0.62} & \multicolumn{1}{l|}{66.38 $\pm$ 5.91} & 77.79 $\pm$ 0.62 & \multicolumn{1}{l|}{77.91 $\pm$ 0.65} & \multicolumn{1}{l|}{66.63 $\pm$ 3.36} & 77.91 $\pm$ 0.65 \\ \hline
\multicolumn{1}{|l|}{\textbf{\textsc{DocEmb} \textsc{FastText} SG}} & \multicolumn{1}{l|}{77.88 $\pm$ 0.01} & \multicolumn{1}{l|}{60.66 $\pm$ 0.01} & 77.88 $\pm$ 0.01 & \multicolumn{1}{l|}{77.88 $\pm$ 0.01} & \multicolumn{1}{l|}{60.66 $\pm$ 0.01} & 77.88 $\pm$ 0.01 \\ \hline
\multicolumn{1}{|l|}{\textbf{\textsc{DocEmb} \textsc{GloVe}}} & \multicolumn{1}{l|}{77.88 $\pm$ 0.01} & \multicolumn{1}{l|}{60.66 $\pm$ 0.01} & 77.88 $\pm$ 0.01 & \multicolumn{1}{l|}{77.88 $\pm$ 0.01} & \multicolumn{1}{l|}{60.66 $\pm$ 0.01} & 77.88 $\pm$ 0.01 \\ \hline
\multicolumn{1}{|l|}{\textbf{\textsc{DocEmb} BERT}} & \multicolumn{1}{l|}{77.76 $\pm$ 1.02} & \multicolumn{1}{l|}{68.20 $\pm$ 3.39} & 77.76 $\pm$ 1.02 & \multicolumn{1}{l|}{77.60 $\pm$ 1.65} & \multicolumn{1}{l|}{70.76 $\pm$ 3.70} & 77.60 $\pm$ 1.65 \\ \hline
\multicolumn{1}{|l|}{\textbf{\textsc{DocEmb} \textsc{RoBERTa}}} & \multicolumn{1}{l|}{77.85 $\pm$ 0.33} & \multicolumn{1}{l|}{63.79 $\pm$ 5.23} & 77.85 $\pm$ 0.33 & \multicolumn{1}{l|}{77.88 $\pm$ 0.01} & \multicolumn{1}{l|}{60.66 $\pm$ 0.01} & 77.88 $\pm$ 0.01 \\ \hline
\multicolumn{1}{|l|}{\textbf{\textsc{DocEmb} BART}} & \multicolumn{1}{l|}{\textbf{79.78 $\pm$ 0.84}} & \multicolumn{1}{l|}{75.84 $\pm$ 1.30} & 79.78 $\pm$ 0.84 & \multicolumn{1}{l|}{79.75 $\pm$ 1.75} & \multicolumn{1}{l|}{75.40 $\pm$ 2.30} & 79.75 $\pm$ 1.75 \\ \hline
 & \multicolumn{3}{|c|}{\textbf{LSTM}} & \multicolumn{3}{c|}{\textbf{Bidirectional LSTM}} \\ \hline
\multicolumn{1}{|l|}{\textbf{Vectorization}} & \multicolumn{1}{c|}{\textbf{Accuracy}} & \multicolumn{1}{c|}{\textbf{Precision}} & \multicolumn{1}{c|}{\textbf{Recall}} & \multicolumn{1}{c|}{\textbf{Accuracy}} & \multicolumn{1}{c|}{\textbf{Precision}} & \multicolumn{1}{c|}{\textbf{Recall}} \\ \hline
\multicolumn{1}{|l|}{\textbf{TFIDF}} & \multicolumn{1}{l|}{78.91 $\pm$ 0.98} & \multicolumn{1}{l|}{73.90 $\pm$ 1.40} & 78.91 $\pm$ 0.98 & \multicolumn{1}{l|}{78.63 $\pm$ 0.80} & \multicolumn{1}{l|}{73.77 $\pm$ 1.63} & 78.63 $\pm$ 0.80 \\ \hline
\multicolumn{1}{|l|}{\textbf{\textsc{DocEmb} \textsc{Word2Vec} CBOW}} & \multicolumn{1}{l|}{77.88 $\pm$ 1.40} & \multicolumn{1}{l|}{70.88 $\pm$ 3.53} & 77.88 $\pm$ 1.40 & \multicolumn{1}{l|}{77.23 $\pm$ 1.55} & \multicolumn{1}{l|}{70.58 $\pm$ 2.30} & 77.23 $\pm$ 1.55 \\ \hline
\multicolumn{1}{|l|}{\textbf{\textsc{DocEmb} \textsc{Word2Vec} SG}} & \multicolumn{1}{l|}{78.04 $\pm$ 0.16} & \multicolumn{1}{l|}{63.35 $\pm$ 1.97} & 78.04 $\pm$ 0.16 & \multicolumn{1}{l|}{77.88 $\pm$ 0.59} & \multicolumn{1}{l|}{67.35 $\pm$ 2.43} & 77.88 $\pm$ 0.59 \\ \hline
\multicolumn{1}{|l|}{\textbf{\textsc{DocEmb} \textsc{FastText} CBOW}} & \multicolumn{1}{l|}{78.07 $\pm$ 0.59} & \multicolumn{1}{l|}{71.74 $\pm$ 2.23} & 78.07 $\pm$ 0.59 & \multicolumn{1}{l|}{78.10 $\pm$ 0.79} & \multicolumn{1}{l|}{73.05 $\pm$ 1.58} & 78.10 $\pm$ 0.79 \\ \hline
\multicolumn{1}{|l|}{\textbf{\textsc{DocEmb} \textsc{FastText} SG}} & \multicolumn{1}{l|}{77.98 $\pm$ 0.20} & \multicolumn{1}{l|}{61.64 $\pm$ 1.69} & 77.98 $\pm$ 0.20 & \multicolumn{1}{l|}{77.85 $\pm$ 0.74} & \multicolumn{1}{l|}{67.51 $\pm$ 5.94} & 77.85 $\pm$ 0.74 \\ \hline
\multicolumn{1}{|l|}{\textbf{\textsc{DocEmb} \textsc{GloVe}}} & \multicolumn{1}{l|}{77.88 $\pm$ 0.01} & \multicolumn{1}{l|}{60.66 $\pm$ 0.01} & 77.88 $\pm$ 0.01 & \multicolumn{1}{l|}{77.85 $\pm$ 0.17} & \multicolumn{1}{l|}{62.74 $\pm$ 3.44} & 77.85 $\pm$ 0.17 \\ \hline
\multicolumn{1}{|l|}{\textbf{\textsc{DocEmb} BERT}} & \multicolumn{1}{l|}{77.57 $\pm$ 2.95} & \multicolumn{1}{l|}{73.20 $\pm$ 2.13} & 77.57 $\pm$ 2.95 & \multicolumn{1}{l|}{77.51 $\pm$ 2.22} & \multicolumn{1}{l|}{74.46 $\pm$ 2.19} & 77.51 $\pm$ 2.22 \\ \hline
\multicolumn{1}{|l|}{\textbf{\textsc{DocEmb} \textsc{RoBERTa}}} & \multicolumn{1}{l|}{77.88 $\pm$ 0.01} & \multicolumn{1}{l|}{60.66 $\pm$ 0.01} & 77.88 $\pm$ 0.01 & \multicolumn{1}{l|}{77.88 $\pm$ 0.01} & \multicolumn{1}{l|}{60.66 $\pm$ 0.01} & 77.88 $\pm$ 0.01 \\ \hline
\multicolumn{1}{|l|}{\textbf{\textsc{DocEmb} BART}} & \multicolumn{1}{l|}{78.07 $\pm$ 2.16} & \multicolumn{1}{l|}{76.31 $\pm$ 3.01} & 78.07 $\pm$ 2.16 & \multicolumn{1}{l|}{77.35 $\pm$ 2.39} & \multicolumn{1}{l|}{75.97 $\pm$ 1.89} & 77.35 $\pm$ 2.39 \\ \hline
 & \multicolumn{3}{|c|}{\textbf{GRU}} & \multicolumn{3}{c|}{\textbf{Bidirectional GRU}} \\ \hline
\multicolumn{1}{|l|}{\textbf{Vectorization}} & \multicolumn{1}{c|}{\textbf{Accuracy}} & \multicolumn{1}{c|}{\textbf{Precision}} & \multicolumn{1}{c|}{\textbf{Recall}} & \multicolumn{1}{c|}{\textbf{Accuracy}} & \multicolumn{1}{c|}{\textbf{Precision}} & \multicolumn{1}{c|}{\textbf{Recall}} \\ \hline
\multicolumn{1}{|l|}{\textbf{TFIDF}} & \multicolumn{1}{l|}{78.94 $\pm$ 1.21} & \multicolumn{1}{l|}{73.99 $\pm$ 2.19} & 78.94 $\pm$ 1.21 & \multicolumn{1}{l|}{78.60 $\pm$ 1.13} & \multicolumn{1}{l|}{73.69 $\pm$ 1.40} & 78.60 $\pm$ 1.13 \\ \hline
\multicolumn{1}{|l|}{\textbf{\textsc{DocEmb} \textsc{Word2Vec} CBOW}} & \multicolumn{1}{l|}{77.57 $\pm$ 1.26} & \multicolumn{1}{l|}{70.72 $\pm$ 2.80} & 77.57 $\pm$ 1.26 & \multicolumn{1}{l|}{77.32 $\pm$ 1.31} & \multicolumn{1}{l|}{71.04 $\pm$ 2.30} & 77.32 $\pm$ 1.31 \\ \hline
\multicolumn{1}{|l|}{\textbf{\textsc{DocEmb} \textsc{Word2Vec} SG}} & \multicolumn{1}{l|}{78.29 $\pm$ 0.37} & \multicolumn{1}{l|}{66.63 $\pm$ 2.52} & 78.29 $\pm$ 0.37 & \multicolumn{1}{l|}{78.22 $\pm$ 1.00} & \multicolumn{1}{l|}{68.24 $\pm$ 2.56} & 78.22 $\pm$ 1.00 \\ \hline
\multicolumn{1}{|l|}{\textbf{\textsc{DocEmb} \textsc{FastText} CBOW}} & \multicolumn{1}{l|}{78.19 $\pm$ 0.78} & \multicolumn{1}{l|}{72.06 $\pm$ 2.37} & 78.19 $\pm$ 0.78 & \multicolumn{1}{l|}{77.73 $\pm$ 1.51} & \multicolumn{1}{l|}{74.22 $\pm$ 1.06} & 77.73 $\pm$ 1.51 \\ \hline
\multicolumn{1}{|l|}{\textbf{\textsc{DocEmb} \textsc{FastText} SG}} & \multicolumn{1}{l|}{77.88 $\pm$ 0.37} & \multicolumn{1}{l|}{64.47 $\pm$ 3.68} & 77.88 $\pm$ 0.37 & \multicolumn{1}{l|}{77.76 $\pm$ 0.79} & \multicolumn{1}{l|}{67.54 $\pm$ 2.82} & 77.76 $\pm$ 0.79 \\ \hline
\multicolumn{1}{|l|}{\textbf{\textsc{DocEmb} \textsc{GloVe}}} & \multicolumn{1}{l|}{77.82 $\pm$ 0.31} & \multicolumn{1}{l|}{61.81 $\pm$ 2.43} & 77.82 $\pm$ 0.31 & \multicolumn{1}{l|}{77.66 $\pm$ 0.52} & \multicolumn{1}{l|}{64.45 $\pm$ 3.17} & 77.66 $\pm$ 0.52 \\ \hline
\multicolumn{1}{|l|}{\textbf{\textsc{DocEmb} BERT}} & \multicolumn{1}{l|}{78.54 $\pm$ 1.39} & \multicolumn{1}{l|}{72.23 $\pm$ 3.39} & 78.54 $\pm$ 1.39 & \multicolumn{1}{l|}{74.86 $\pm$ 4.28} & \multicolumn{1}{l|}{72.99 $\pm$ 2.21} & 74.86 $\pm$ 4.28 \\ \hline
\multicolumn{1}{|l|}{\textbf{\textsc{DocEmb} \textsc{RoBERTa}}} & \multicolumn{1}{l|}{77.88 $\pm$ 0.01} & \multicolumn{1}{l|}{60.66 $\pm$ 0.01} & 77.88 $\pm$ 0.01 & \multicolumn{1}{l|}{77.88 $\pm$ 0.01} & \multicolumn{1}{l|}{60.66 $\pm$ 0.01} & 77.88 $\pm$ 0.01 \\ \hline
\multicolumn{1}{|l|}{\textbf{\textsc{DocEmb} BART}} & \multicolumn{1}{l|}{77.48 $\pm$ 2.08} & \multicolumn{1}{l|}{75.96 $\pm$ 2.38} & 77.48 $\pm$ 2.08 & \multicolumn{1}{l|}{76.45 $\pm$ 4.56} & \multicolumn{1}{l|}{77.31 $\pm$ 3.14} & 76.45 $\pm$ 4.56 \\ \hline
& \multicolumn{2}{c|}{\textbf{Accuracy}} & \multicolumn{2}{c|}{\textbf{Precision}} & \multicolumn{2}{c|}{\textbf{Recall}} \\ \hline
\multicolumn{1}{|l|}{\textbf{MisRoBÆRTa~\citep{Truica2022}}}   & \multicolumn{2}{c|}{77.39 $\pm$ 0.83}         & \multicolumn{2}{c|}{77.39 $\pm$ 0.83}          & \multicolumn{2}{c|}{77.39 $\pm$ 0.83}       \\ \hline
& & \multicolumn{5}{|c|}{\textbf{Accuracy}}  \\ \hline
\multicolumn{2}{|l|}{\textbf{SVM~\citep{Horne2017}}}   & \multicolumn{5}{c|}{78.00}        \\ \hline
\multicolumn{2}{|l|}{\textbf{UFD~\citep{Yang2019}}}    & \multicolumn{5}{c|}{67.90}        \\ \hline
\end{tabular}
}
\end{table*}

\begin{table*}[!htb]
\centering
\caption{Fake news detection results on TSHP-17 dataset as presented in~\citet{Rashkin2017} and ~\citet{BarronCedeno2019a}}
\label{tab:results_polifact}
\resizebox{1\textwidth}{!}{%
\begin{tabular}{llll|lll|}
\cline{2-7}
 & \multicolumn{3}{|c|}{\textbf{Naïve Bayes}} & \multicolumn{3}{c|}{\textbf{Gradient   Boosted Trees}} \\ \hline
\multicolumn{1}{|l|}{\textbf{Vectorization}} & \multicolumn{1}{c|}{\textbf{Accuracy}} & \multicolumn{1}{c|}{\textbf{Precision}} & \multicolumn{1}{c|}{\textbf{Recall}} & \multicolumn{1}{c|}{\textbf{Accuracy}} & \multicolumn{1}{c|}{\textbf{Precision}} & \multicolumn{1}{c|}{\textbf{Recall}} \\ \hline
\multicolumn{1}{|l|}{\textbf{TFIDF}} & \multicolumn{1}{l|}{92.08 $\pm$ 0.15} & \multicolumn{1}{l|}{92.11 $\pm$ 0.15} & 92.08 $\pm$ 0.15 & \multicolumn{1}{l|}{98.05 $\pm$ 0.12} & \multicolumn{1}{l|}{98.05 $\pm$ 0.12} & 98.05 $\pm$ 0.12 \\ \hline
\multicolumn{1}{|l|}{\textbf{\textsc{DocEmb} \textsc{Word2Vec} CBOW}} & \multicolumn{1}{l|}{70.06 $\pm$ 0.50} & \multicolumn{1}{l|}{73.01 $\pm$ 0.36} & 70.06 $\pm$ 0.50 & \multicolumn{1}{l|}{95.64 $\pm$ 0.26} & \multicolumn{1}{l|}{95.63 $\pm$ 0.26} & 95.64 $\pm$ 0.26 \\ \hline
\multicolumn{1}{|l|}{\textbf{\textsc{DocEmb} \textsc{Word2Vec} SG}} & \multicolumn{1}{l|}{55.33 $\pm$ 0.63} & \multicolumn{1}{l|}{68.18 $\pm$ 0.33} & 55.33 $\pm$ 0.63 & \multicolumn{1}{l|}{95.76 $\pm$ 0.21} & \multicolumn{1}{l|}{95.75 $\pm$ 0.21} & 95.76 $\pm$ 0.21 \\ \hline
\multicolumn{1}{|l|}{\textbf{\textsc{DocEmb} \textsc{FastText} CBOW}} & \multicolumn{1}{l|}{62.83 $\pm$ 0.47} & \multicolumn{1}{l|}{70.17 $\pm$ 0.66} & 62.83 $\pm$ 0.47 & \multicolumn{1}{l|}{94.36 $\pm$ 0.27} & \multicolumn{1}{l|}{94.34 $\pm$ 0.27} & 94.36 $\pm$ 0.27 \\ \hline
\multicolumn{1}{|l|}{\textbf{\textsc{DocEmb} \textsc{FastText} SG}} & \multicolumn{1}{l|}{59.72 $\pm$ 0.52} & \multicolumn{1}{l|}{69.49 $\pm$ 0.63} & 59.72 $\pm$ 0.52 & \multicolumn{1}{l|}{95.66 $\pm$ 0.28} & \multicolumn{1}{l|}{95.65 $\pm$ 0.28} & 95.66 $\pm$ 0.28 \\ \hline
\multicolumn{1}{|l|}{\textbf{\textsc{DocEmb} \textsc{GloVe}}} & \multicolumn{1}{l|}{52.99 $\pm$ 0.56} & \multicolumn{1}{l|}{65.84 $\pm$ 0.43} & 52.99 $\pm$ 0.56 & \multicolumn{1}{l|}{96.19 $\pm$ 0.28} & \multicolumn{1}{l|}{96.19 $\pm$ 0.28} & 96.19 $\pm$ 0.28 \\ \hline
\multicolumn{1}{|l|}{\textbf{\textsc{DocEmb} BERT}} & \multicolumn{1}{l|}{84.65 $\pm$ 0.63} & \multicolumn{1}{l|}{87.60 $\pm$ 0.42} & 84.65 $\pm$ 0.63 & \multicolumn{1}{l|}{98.18 $\pm$ 0.10} & \multicolumn{1}{l|}{98.18 $\pm$ 0.10} & 98.18 $\pm$ 0.10 \\ \hline
\multicolumn{1}{|l|}{\textbf{\textsc{DocEmb} \textsc{RoBERTa}}} & \multicolumn{1}{l|}{52.15 $\pm$ 0.33} & \multicolumn{1}{l|}{65.77 $\pm$ 0.64} & 52.15 $\pm$ 0.33 & \multicolumn{1}{l|}{79.15 $\pm$ 0.36} & \multicolumn{1}{l|}{79.11 $\pm$ 0.38} & 79.15 $\pm$ 0.36 \\ \hline
\multicolumn{1}{|l|}{\textbf{\textsc{DocEmb} BART}} & \multicolumn{1}{l|}{94.08 $\pm$ 0.28} & \multicolumn{1}{l|}{94.66 $\pm$ 0.26} & 94.08 $\pm$ 0.28 & \multicolumn{1}{l|}{99.01 $\pm$ 0.10} & \multicolumn{1}{l|}{99.01 $\pm$ 0.10} & 99.01 $\pm$ 0.10 \\ \hline
\textbf{} & \multicolumn{3}{|c|}{\textbf{Perceptron}} & \multicolumn{3}{c|}{\textbf{Multi Layer Perceptron}} \\ \hline
\multicolumn{1}{|l|}{\textbf{Vectorization}} & \multicolumn{1}{c|}{\textbf{Accuracy}} & \multicolumn{1}{c|}{\textbf{Precision}} & \multicolumn{1}{c|}{\textbf{Recall}} & \multicolumn{1}{c|}{\textbf{Accuracy}} & \multicolumn{1}{c|}{\textbf{Precision}} & \multicolumn{1}{c|}{\textbf{Recall}} \\ \hline
\multicolumn{1}{|l|}{\textbf{TFIDF}} & \multicolumn{1}{l|}{97.64 $\pm$ 0.13} & \multicolumn{1}{l|}{97.64 $\pm$ 0.13} & 97.64 $\pm$ 0.13 & \multicolumn{1}{l|}{97.54 $\pm$ 0.16} & \multicolumn{1}{l|}{97.54 $\pm$ 0.16} & 97.54 $\pm$ 0.16 \\ \hline
\multicolumn{1}{|l|}{\textbf{\textsc{DocEmb} \textsc{Word2Vec} CBOW}} & \multicolumn{1}{l|}{94.24 $\pm$ 0.23} & \multicolumn{1}{l|}{94.22 $\pm$ 0.24} & 94.24 $\pm$ 0.23 & \multicolumn{1}{l|}{96.11 $\pm$ 0.22} & \multicolumn{1}{l|}{96.11 $\pm$ 0.22} & 96.11 $\pm$ 0.22 \\ \hline
\multicolumn{1}{|l|}{\textbf{\textsc{DocEmb} \textsc{Word2Vec} SG}} & \multicolumn{1}{l|}{90.14 $\pm$ 0.29} & \multicolumn{1}{l|}{90.09 $\pm$ 0.29} & 90.14 $\pm$ 0.29 & \multicolumn{1}{l|}{93.91 $\pm$ 0.24} & \multicolumn{1}{l|}{93.89 $\pm$ 0.24} & 93.91 $\pm$ 0.24 \\ \hline
\multicolumn{1}{|l|}{\textbf{\textsc{DocEmb} \textsc{FastText} CBOW}} & \multicolumn{1}{l|}{93.14 $\pm$ 0.30} & \multicolumn{1}{l|}{93.14 $\pm$ 0.30} & 93.14 $\pm$ 0.30 & \multicolumn{1}{l|}{95.63 $\pm$ 0.24} & \multicolumn{1}{l|}{95.64 $\pm$ 0.24} & 95.63 $\pm$ 0.24 \\ \hline
\multicolumn{1}{|l|}{\textbf{\textsc{DocEmb} \textsc{FastText} SG}} & \multicolumn{1}{l|}{90.00 $\pm$ 0.20} & \multicolumn{1}{l|}{89.94 $\pm$ 0.21} & 90.00 $\pm$ 0.20 & \multicolumn{1}{l|}{93.64 $\pm$ 0.30} & \multicolumn{1}{l|}{93.64 $\pm$ 0.29} & 93.64 $\pm$ 0.30 \\ \hline
\multicolumn{1}{|l|}{\textbf{\textsc{DocEmb} \textsc{GloVe}}} & \multicolumn{1}{l|}{90.97 $\pm$ 0.27} & \multicolumn{1}{l|}{90.95 $\pm$ 0.27} & 90.97 $\pm$ 0.27 & \multicolumn{1}{l|}{94.04 $\pm$ 0.30} & \multicolumn{1}{l|}{94.05 $\pm$ 0.29} & 94.04 $\pm$ 0.30 \\ \hline
\multicolumn{1}{|l|}{\textbf{\textsc{DocEmb} BERT}} & \multicolumn{1}{l|}{98.44 $\pm$ 0.15} & \multicolumn{1}{l|}{98.44 $\pm$ 0.15} & 98.44 $\pm$ 0.15 & \multicolumn{1}{l|}{98.78 $\pm$ 0.14} & \multicolumn{1}{l|}{98.78 $\pm$ 0.13} & 98.78 $\pm$ 0.14 \\ \hline
\multicolumn{1}{|l|}{\textbf{\textsc{DocEmb} \textsc{RoBERTa}}} & \multicolumn{1}{l|}{77.79 $\pm$ 1.85} & \multicolumn{1}{l|}{78.89 $\pm$ 0.65} & 77.79 $\pm$ 1.85 & \multicolumn{1}{l|}{80.30 $\pm$ 1.55} & \multicolumn{1}{l|}{81.29 $\pm$ 0.61} & 80.30 $\pm$ 1.55 \\ \hline
\multicolumn{1}{|l|}{\textbf{\textsc{DocEmb} BART}} & \multicolumn{1}{l|}{99.54 $\pm$ 0.05} & \multicolumn{1}{l|}{99.54 $\pm$ 0.05} & 99.54 $\pm$ 0.05 & \multicolumn{1}{l|}{99.55 $\pm$ 0.06} & \multicolumn{1}{l|}{99.55 $\pm$ 0.06} & 99.55 $\pm$ 0.06 \\ \hline
 & \multicolumn{3}{|c|}{\textbf{LSTM}} & \multicolumn{3}{c|}{\textbf{Bidirectional LSTM}} \\ \hline
\multicolumn{1}{|l|}{\textbf{Vectorization}} & \multicolumn{1}{c|}{\textbf{Accuracy}} & \multicolumn{1}{c|}{\textbf{Precision}} & \multicolumn{1}{c|}{\textbf{Recall}} & \multicolumn{1}{c|}{\textbf{Accuracy}} & \multicolumn{1}{c|}{\textbf{Precision}} & \multicolumn{1}{c|}{\textbf{Recall}} \\ \hline
\multicolumn{1}{|l|}{\textbf{TFIDF}} & \multicolumn{1}{l|}{97.10 $\pm$ 0.17} & \multicolumn{1}{l|}{97.10 $\pm$ 0.17} & 97.10 $\pm$ 0.17 & \multicolumn{1}{l|}{97.03 $\pm$ 0.22} & \multicolumn{1}{l|}{97.03 $\pm$ 0.22} & 97.03 $\pm$ 0.22 \\ \hline
\multicolumn{1}{|l|}{\textbf{\textsc{DocEmb} \textsc{Word2Vec} CBOW}} & \multicolumn{1}{l|}{96.93 $\pm$ 0.16} & \multicolumn{1}{l|}{96.94 $\pm$ 0.15} & 96.93 $\pm$ 0.16 & \multicolumn{1}{l|}{96.88 $\pm$ 0.09} & \multicolumn{1}{l|}{96.88 $\pm$ 0.09} & 96.88 $\pm$ 0.09 \\ \hline
\multicolumn{1}{|l|}{\textbf{\textsc{DocEmb} \textsc{Word2Vec} SG}} & \multicolumn{1}{l|}{95.37 $\pm$ 0.17} & \multicolumn{1}{l|}{95.40 $\pm$ 0.17} & 95.37 $\pm$ 0.17 & \multicolumn{1}{l|}{95.55 $\pm$ 0.32} & \multicolumn{1}{l|}{95.56 $\pm$ 0.30} & 95.55 $\pm$ 0.32 \\ \hline
\multicolumn{1}{|l|}{\textbf{\textsc{DocEmb} \textsc{FastText} CBOW}} & \multicolumn{1}{l|}{96.24 $\pm$ 0.30} & \multicolumn{1}{l|}{96.26 $\pm$ 0.29} & 96.24 $\pm$ 0.30 & \multicolumn{1}{l|}{96.37 $\pm$ 0.23} & \multicolumn{1}{l|}{96.38 $\pm$ 0.23} & 96.37 $\pm$ 0.23 \\ \hline
\multicolumn{1}{|l|}{\textbf{\textsc{DocEmb} \textsc{FastText} SG}} & \multicolumn{1}{l|}{95.06 $\pm$ 0.13} & \multicolumn{1}{l|}{95.07 $\pm$ 0.13} & 95.06 $\pm$ 0.13 & \multicolumn{1}{l|}{95.10 $\pm$ 0.37} & \multicolumn{1}{l|}{95.11 $\pm$ 0.32} & 95.10 $\pm$ 0.37 \\ \hline
\multicolumn{1}{|l|}{\textbf{\textsc{DocEmb} \textsc{GloVe}}} & \multicolumn{1}{l|}{95.17 $\pm$ 0.34} & \multicolumn{1}{l|}{95.22 $\pm$ 0.28} & 95.17 $\pm$ 0.34 & \multicolumn{1}{l|}{95.31 $\pm$ 0.42} & \multicolumn{1}{l|}{95.35 $\pm$ 0.36} & 95.31 $\pm$ 0.42 \\ \hline
\multicolumn{1}{|l|}{\textbf{\textsc{DocEmb} BERT}} & \multicolumn{1}{l|}{98.86 $\pm$ 0.24} & \multicolumn{1}{l|}{98.87 $\pm$ 0.22} & 98.86 $\pm$ 0.24 & \multicolumn{1}{l|}{98.89 $\pm$ 0.17} & \multicolumn{1}{l|}{98.89 $\pm$ 0.16} & 98.89 $\pm$ 0.17 \\ \hline
\multicolumn{1}{|l|}{\textbf{\textsc{DocEmb} \textsc{RoBERTa}}} & \multicolumn{1}{l|}{80.25 $\pm$ 1.38} & \multicolumn{1}{l|}{81.31 $\pm$ 0.73} & 80.25 $\pm$ 1.38 & \multicolumn{1}{l|}{80.17 $\pm$ 1.66} & \multicolumn{1}{l|}{81.30 $\pm$ 0.86} & 80.17 $\pm$ 1.66 \\ \hline
\multicolumn{1}{|l|}{\textbf{\textsc{DocEmb} BART}} & \multicolumn{1}{l|}{99.62 $\pm$ 0.05} & \multicolumn{1}{l|}{99.62 $\pm$ 0.05} & 99.62 $\pm$ 0.05 & \multicolumn{1}{l|}{\textbf{99.65 $\pm$ 0.05}} & \multicolumn{1}{l|}{99.65 $\pm$ 0.05} & 99.65 $\pm$ 0.05 \\ \hline
 & \multicolumn{3}{|c|}{\textbf{GRU}} & \multicolumn{3}{c|}{\textbf{Bidirectional GRU}} \\ \hline
\multicolumn{1}{|l|}{\textbf{Vectorization}} & \multicolumn{1}{c|}{\textbf{Accuracy}} & \multicolumn{1}{c|}{\textbf{Precision}} & \multicolumn{1}{c|}{\textbf{Recall}} & \multicolumn{1}{c|}{\textbf{Accuracy}} & \multicolumn{1}{c|}{\textbf{Precision}} & \multicolumn{1}{c|}{\textbf{Recall}} \\ \hline
\multicolumn{1}{|l|}{\textbf{TFIDF}} & \multicolumn{1}{l|}{97.00 $\pm$ 0.13} & \multicolumn{1}{l|}{97.00 $\pm$ 0.13} & 97.00 $\pm$ 0.13 & \multicolumn{1}{l|}{96.85 $\pm$ 0.20} & \multicolumn{1}{l|}{96.86 $\pm$ 0.20} & 96.85 $\pm$ 0.20 \\ \hline
\multicolumn{1}{|l|}{\textbf{\textsc{DocEmb} \textsc{Word2Vec} CBOW}} & \multicolumn{1}{l|}{96.85 $\pm$ 0.16} & \multicolumn{1}{l|}{96.86 $\pm$ 0.16} & 96.85 $\pm$ 0.16 & \multicolumn{1}{l|}{96.86 $\pm$ 0.14} & \multicolumn{1}{l|}{96.87 $\pm$ 0.14} & 96.86 $\pm$ 0.14 \\ \hline
\multicolumn{1}{|l|}{\textbf{\textsc{DocEmb} \textsc{Word2Vec} SG}} & \multicolumn{1}{l|}{95.29 $\pm$ 0.28} & \multicolumn{1}{l|}{95.31 $\pm$ 0.26} & 95.29 $\pm$ 0.28 & \multicolumn{1}{l|}{95.53 $\pm$ 0.14} & \multicolumn{1}{l|}{95.56 $\pm$ 0.14} & 95.53 $\pm$ 0.14 \\ \hline
\multicolumn{1}{|l|}{\textbf{\textsc{DocEmb} \textsc{FastText} CBOW}} & \multicolumn{1}{l|}{96.28 $\pm$ 0.17} & \multicolumn{1}{l|}{96.29 $\pm$ 0.17} & 96.28 $\pm$ 0.17 & \multicolumn{1}{l|}{96.25 $\pm$ 0.28} & \multicolumn{1}{l|}{96.26 $\pm$ 0.26} & 96.25 $\pm$ 0.28 \\ \hline
\multicolumn{1}{|l|}{\textbf{\textsc{DocEmb} \textsc{FastText} SG}} & \multicolumn{1}{l|}{94.72 $\pm$ 0.31} & \multicolumn{1}{l|}{94.75 $\pm$ 0.28} & 94.72 $\pm$ 0.31 & \multicolumn{1}{l|}{94.77 $\pm$ 0.52} & \multicolumn{1}{l|}{94.85 $\pm$ 0.44} & 94.77 $\pm$ 0.52 \\ \hline
\multicolumn{1}{|l|}{\textbf{\textsc{DocEmb} \textsc{GloVe}}} & \multicolumn{1}{l|}{95.06 $\pm$ 0.34} & \multicolumn{1}{l|}{95.10 $\pm$ 0.31} & 95.06 $\pm$ 0.34 & \multicolumn{1}{l|}{95.09 $\pm$ 0.29} & \multicolumn{1}{l|}{95.15 $\pm$ 0.25} & 95.09 $\pm$ 0.29 \\ \hline
\multicolumn{1}{|l|}{\textbf{\textsc{DocEmb} BERT}} & \multicolumn{1}{l|}{98.91 $\pm$ 0.27} & \multicolumn{1}{l|}{98.92 $\pm$ 0.25} & 98.91 $\pm$ 0.27 & \multicolumn{1}{l|}{98.99 $\pm$ 0.10} & \multicolumn{1}{l|}{98.99 $\pm$ 0.10} & 98.99 $\pm$ 0.10 \\ \hline
\multicolumn{1}{|l|}{\textbf{\textsc{DocEmb} \textsc{RoBERTa}}} & \multicolumn{1}{l|}{80.44 $\pm$ 1.26} & \multicolumn{1}{l|}{81.44 $\pm$ 0.55} & 80.44 $\pm$ 1.26 & \multicolumn{1}{l|}{80.04 $\pm$ 1.69} & \multicolumn{1}{l|}{81.36 $\pm$ 0.76} & 80.04 $\pm$ 1.69 \\ \hline
\multicolumn{1}{|l|}{\textbf{\textsc{DocEmb} BART}} & \multicolumn{1}{l|}{99.62 $\pm$ 0.09} & \multicolumn{1}{l|}{99.62 $\pm$ 0.09} & 99.62 $\pm$ 0.09 & \multicolumn{1}{l|}{99.64 $\pm$ 0.07} & \multicolumn{1}{l|}{99.64 $\pm$ 0.07} & 99.64 $\pm$ 0.07 \\ \hline
& \multicolumn{2}{c|}{\textbf{Accuracy}} & \multicolumn{2}{c|}{\textbf{Precision}} & \multicolumn{2}{c|}{\textbf{Recall}} \\ \hline
\multicolumn{1}{|l|}{\textbf{MisRoBÆRTa~\citep{Truica2022}}}   & \multicolumn{2}{c|}{99.52 $\pm$ 0.12}         & \multicolumn{2}{c|}{99.52 $\pm$ 0.12}          & \multicolumn{2}{c|}{99.52 $\pm$ 0.12}       \\ \hline

& & \multicolumn{5}{|c|}{\textbf{Accuracy}}  \\ \hline
\multicolumn{2}{|l|}{\textbf{Proppy~\citep{BarronCedeno2019b}}}   & \multicolumn{5}{c|}{98.36}        \\ \hline
\end{tabular}
}
\end{table*}

Table~\ref{tab:results_kagle_full} presents the results obtained on the Kaggle dataset~\citep{Kaliyar2020,Kaliyar2021}.
We observed that only for the Gradient Boosted Trees and Multi-Layer Perceptron models, the document embeddings obtained with \textsc{Word2Vec} SG and \textsc{FastText} SG outperformed their CBOW counterparts.
When analyzing the same document embedding, i.e., \textsc{DocEmb}, we observed very little difference in performance among the neural models.
As we used early stopping mechanisms, the neural network models did not overfit.
Among the document embeddings employing transformers, the ones that use BART obtained the best results across all experiments.
With an accuracy of $99.80\%$, the overall best-performing model is the Bidirectional LSTM with document embeddings constructed with BART, i.e., \textsc{DocEmb BART}.
The results show that our approach outperforms more complex models proposed in the current literature, e.g., 
FNDNet~\citep{Kaliyar2020} obtained an accuracy of $98.36\%$, FakeBERT~\citep{Kaliyar2021} obtained an accuracy of $98.90\%$, and C-CNN~\citep{Sedik2022} obtained an accuracy of $99.90\%$.
We observe that C-CNN, a large neural model with multiple layers that also concatenates the results of three CNN models, outperforms  the Bidirectional LSTM in terms of average accuracy on the Kaggle dataset by only $0.10\%$.
We also want to emphasize that the results in Table~\ref{tab:results_kagle_full} present the mean over 10 runs for each metric per model and embedding pair.
Thus, if we only take the best-performing model {as in the case of the C-CNN results presented by}~\citet{Sedik2022}, then the Bidirectional LSTM model manages to obtain an accuracy of $99.92\% (= 99.80\%$ (mean accuracy) $+ 0.12\%$ (standard deviation)).

Table~\ref{tab:results_buzzfeed} presents the results obtained on the Buzz Feed News dataset~\citep{Horne2017}.
On this dataset, we observed that all the models obtained good results with TFIDF, such that some models that employ the TFIDF vectorization outperformed the document embeddings constructed with word and transformer embeddings, see, e.g., the results for LSTM, Bidirectional LSTM, GRU, and Bidirectional GRU.
With an accuracy of $79.78\%$, the overall best-performing model is  Perceptron with BART document embeddings.
For all the models, there is very little difference between the document embeddings that employ CBOW and their Skip-Gram counterparts.
The results show that our approach outperforms more complex models proposed in the current literature, e.g., 
SVM~\citep{Horne2017} obtained an accuracy of $78.00\%$ and UFD~\citep{Yang2019} obtained an accuracy of $67.90\%$.

Table~\ref{tab:results_polifact} presents our final experiments, on the TSHP-17 dataset~\citep{Rashkin2017,BarronCedeno2019a}.
We observed that the document embeddings (\textsc{DocEmb}s) obtained with \textsc{Word2Vec} SG and \textsc{FastText} SG outperformed their CBOW counterparts only for the Gradient Boosted Trees model.
All the models that employ TFIDF outperformed their counterparts that employ document embedding built with word embeddings. 
Among the document embeddings employing a transformer, the ones built with BART obtained the best results with regards to the model, while the ones that employ \textsc{RoBERTa} obtained the worst results.
With an accuracy of $99.65\%$, the overall best-performing model is the Bidirectional LSTM with BART document embeddings.
For the same \textsc{DocEmb}, we observed very little difference in performance among the neural models. 
The results show that our approach outperforms more complex models proposed in the current literature, e.g., Proppy's~\citep{BarronCedeno2019b} accuracy is $98.36\%$.

Thus, in conclusion, we show, on five additional datasets, that:
\begin{itemize}
    \item[(1)] A simpler neural architecture offers at least similar or better results as complex architectures that employ multiple layers, and
    \item[(2)] The difference in performance lies in the embeddings used to vectorize the textual data.
\end{itemize}

Furthermore, we {generally} obtained better results than in other current state-of-the-art work:
\begin{itemize}
    \item[(1)] On the LIAR dataset with 6 labels, \citet{Wang2017} obtained an $F_1$-Score of $27.7\%$ using Hybrid CNNs and \citet{Alhindi2018} obtained an $F_1$-Score of $26\%$ using BiLSTM, while we obtained an accuracy of $25.89\%$ using Multi-Layer Perceptron with the document embeddings employing BART;
    \item[(2)] On the LIAR dataset with 2 labels, \citet{Upadhayay2020} obtained an accuracy of $70\%$ using CNN with BERT-base embeddings, while we obtained an accuracy of $83.99\%$ using LSTM with the document embeddings employing \textsc{GloVe};
    \item[(3)] On the Kaggle dataset, the large deep learning model FakeBERT~\citep{Kaliyar2021} obtained an accuracy of $98.90\%$ and C-CNN~\citep{Sedik2022} obtained an accuracy of $99.90\%$, while we obtained an accuracy of $99.80\%$ using a simple Bidirectional LSTM with the document embeddings employing BART;
    \item[(4)] On the Buzz Feed News dataset, \citet{Horne2017} obtained an accuracy of $78\%$ using a linear SVM, while we obtained an accuracy of $79.78\%$ using Perceptron with the document embeddings employing BART;
    \item[(5)] On the TSHP-17 dataset, \citet{BarronCedeno2019a} obtained an accuracy of $97.58\%$ using Proppy~\citep{BarronCedeno2019b}, while we obtained $99.65\%$ using Bidirectional LSTM with the document embeddings employing BART;
\end{itemize}

To sum up, this set of experiments again enforces our observations that the embedding is more important than the complexity of the classification architecture.
Furthermore, there is no generic model that offers the best performance regardless of the dataset.
Thus, a data-driven approach together with hyper-parameter tuning and ablation testing should be considered when the goal is to determine the best-performing model for a given dataset.

\section{Discussion}\label{sec:discussion}

Word embeddings manage to capture both local and global contexts as defined in~\citet{Truica2021a}.
These help the machine learning algorithms to model and learn the text context, syntax, and semantics, but fail to differentiate among the words' grammatical functions, i.e., the same word embedding is computed for a word regardless of its part-of-speech.
On the other hand, transformer embeddings manage to learn the linguistic meaning of words, as they manage to preserve context by design.
Thus, the same word has a different embedding depending on its lexical sense and concept as well as part-of-speech.
Based on this, we can observe that the experiments that use document embeddings that employ transformers perform better than those that employ word embedding  on average.
The most interesting results, however, are those obtained with the document representation obtained with TFIDF.
We observed that only the frequency-based importance of a word to a document within a textual corpus has a high impact on the models' performance.
As a general observation, we observed very little difference in performance among the neural models when using the same document embedding.

The experimental results show that the \textsc{DocEmb}s that use \textsc{Word2Vec} and \textsc{FastText} obtain very similar results, with a difference of $\sim$$\pm$2\% when using Perceptron, and $\sim$$\pm$0.5\% when using LSTM.
\textsc{GloVe} based \textsc{DocEmb} obtains the best results together with the LSTM model on the LIAR dataset when using 2 labels.
For the  sample extracted from the Fake News Corpus as well as the LIAR with 6 labels, Kaggle, and TSHP-17 datasets, the BART \textsc{DocEmb} obtains the best results with different classification algorithms.
We can conclude that there is no clear result for a classification model that generalizes well regardless of the dataset. 

From the experimental results, we could not determine a clear winner with regards to document embedding and classification model.
We observed empirically that the best-performing classification model changes with the dataset and the document embedding employed.

Our \textsc{DocEmb} solutions were compared to the results we obtained when employing MisRoBÆRTa~\citep{Truica2022}, a more complex state-of-the-art model that employs fine-tune BART and \textsc{RoBERTa} embeddings.
We note that we did not use fine-tuning for our dataset as the authors did in the original work~\citep{Truica2022}.
Thus, we used the pre-trained BART (\textit{facebook/bart-large}) and \textsc{RoBERTa} (\textit{roberta-base}) from HuggingFage~\citep{Wolf2020}.
Furthermore, we also compared the results we obtained on each dataset with the results obtained with other state-of-the-art models presented in the current literature.

In our experiments, we obtained results that lead to the following observation: feature selection is more important than the Deep Learning Architecture used for classification.
To put it bluntly, the need to stack layers upon layers of neural cells, just to claim a novel architecture, does not solve real-world problems, it just exacerbates out of proportion our understanding of how to use Machine Learning/Deep Learning for Natural Language Processing tasks.
To conclude our findings:
\begin{itemize}
    \item[(1)] A simpler neural architecture offers similar if not better results as complex deep learning architectures that employ multiple layers, i.e., in our comparison, we obtained similar results as the complex MisRoBÆRTa~\citep{Truica2022} architecture, better than state-of-the-art results, i.e., FakeBERT~\citep{Kaliyar2021}, and Poppy~\citep{BarronCedeno2019b};
    \item[(2)] The embeddings used to vectorize the textual data makes all the difference in performance, i.e., the right embedding must be selected to obtain good results with a given model; 
    \item[(3)] We need a data-driven approach to select the best model and the best embedding for our dataset;
    \item[(4)] The way the word embedding manages to encapsulate the semantic, syntactic, and context features improves the performance of the classification models.
\end{itemize}

\section{Conclusions}\label{sec:conclusions}

In this article, we presented a new approach for fake news detection using document embeddings (\textsc{DocEmb}s).
We also proposed a benchmark to establish the most efficient ways for finding misleading information.
To detect fake news, we used multiple machine learning algorithms together with \textsc{DocEmb}s built using either TFIDF, or word and transformer embeddings: \textsc{Word2Vec} SG and CBOW, \textsc{FastText} SG and CBOW, \textsc{GloVe}, BERT, \textsc{RoBERTa}, and BART.

Our approach emphasizes the importance of an overall document representation when dealing with the task of fake news detection and shows state-of-the-art performance results.
Depending on the dataset, the results show that \textsc{BiGRU}/\textsc{BiLSTM} with \textsc{DocEmb} BART outperforms the other models.
In the experiments, we obtained better results than state-of-the-art Deep Neural Network models,
even though we used a simpler Deep Neural Network Architecture.
Additionally, we obtained similar results as MisRoBÆRTa~\citep{Truica2022} when using pre-trained BART (\textit{facebook/bart-large}) and \textsc{RoBERTa} (\textit{roberta-base}) from HuggingFace~\citep{Wolf2020}.
These are significant results, not because of the evaluation scores but because of the complexity of the models.
The main takeaway of this work is that a simpler neural architecture offers similar if not better results as complex architectures that employ multiple layers. 
We observe that the most relevant factor is the embedding employed for classification, as it can really make a difference.

In future research, we plan to use sentiment analysis with fake news detection to determine if there is a correlation between polarity and veracity.
We also aim to use ensemble models that combine our proposed method with existing methods to determine if the performance of fake news detection is improved.

\paragraph*{Acknowledgement:} This work was done within the ``AI-based conversational agent for misinformation fact-checking'' project financed through the OPTIM Research framework (POCU grant no. 62461/03.06.2022, SMIS code 153735) and partially funded by the University Politehnica of Bucharest through the PubArt program.

\end{document}